\documentclass[10pt,twocolumn,letterpaper]{article}

\usepackage{iccv}
\usepackage{times}
\usepackage[accsupp]{axessibility}
\usepackage[position=top]{subfig}

\usepackage{cite}
\usepackage[utf8]{inputenc}
\usepackage{graphicx}
\usepackage{amsthm}
\usepackage{dsfont}
\usepackage{amssymb}
\usepackage[shortlabels]{enumitem}
\usepackage{graphicx}
\usepackage{mathrsfs}
\usepackage{float}
\usepackage{bbm}
\usepackage{bm}
\usepackage{amsmath}
\usepackage{comment}
\usepackage{listings}
\usepackage{color}
\usepackage{booktabs}
\usepackage{adjustbox}
\usepackage{hhline}

\usepackage{multicol}
\usepackage{multirow}
\usepackage{soul}

\usepackage{algorithm}
\usepackage{algorithmicx}
\usepackage{algpseudocode}
\allowdisplaybreaks

\graphicspath{{eps/}{video/}}

\newtheorem{theorem}{Theorem}[section]

\newtheorem{definition}[theorem]{Definition}

\theoremstyle{definition}

\newtheorem{remark}[theorem]{Remark}

\DeclareMathOperator*{\minimize}{\mathrm{minimize}}
\DeclareMathOperator*{\subject}{\mathrm{subject~to~}}

\def\BL{\bm{L}}

\def\BS{\bm{S}}

\def\BX{\bm{X}}

\def\cE{\mathcal{E}}
\def\cR{\mathcal{R}}
\def\cX{\mathcal{X}}
\def\cL{\mathcal{L}}
\def\cS{\mathcal{S}}

\def\cO{\mathcal{O}}
\def\fro{\mathrm{F}}
\newcommand{\tens}[1]{\mathcal{#1}}

\newcommand{\alg}{RTCUR}
\newcommand{\algf}{\alg-F}
\newcommand{\algr}{\alg-R}

\newcommand{\supp}{{\rm supp\,}}
\newcommand{\rank}{{\rm rank\,}}

\definecolor{OliveGreen}{rgb}{0.23,0.39,0.30}

\definecolor{dncolor}{rgb}{0.85, 0.15, 0.65}

\usepackage[breaklinks=true,bookmarks=false]{hyperref}
\hypersetup{
	colorlinks,
	citecolor=blue,
	filecolor=blue,
	linkcolor=blue,
	urlcolor=blue,
	hyperfootnotes=false
}

\iccvfinalcopy 

\def\iccvPaperID{****} 

\begin{document}

\title{Fast Robust Tensor Principal Component Analysis \\via Fiber CUR Decomposition\thanks{This work was supported in part by grants NSF BIGDATA DMS \#1740325 and NSF DMS \#2011140. }
}

\author{HanQin Cai \qquad Zehan Chao \qquad Longxiu Huang \qquad Deanna Needell\\
Department of Mathematics \protect\\ University of California, Los Angeles\protect\\ Los Angeles, CA 90095, USA\\
{\tt\small \{hqcai,zchao,huangl3,deanna\}@math.ucla.edu }
}

\maketitle

\begin{abstract}
    We study the problem of tensor robust principal component analysis (TRPCA), which aims to separate an underlying low-multilinear-rank tensor and a sparse outlier tensor from their sum. In this work, we propose a fast non-convex algorithm, coined Robust Tensor CUR (\alg), for large-scale TRPCA problems. \alg\ considers a framework of alternating projections and utilizes the recently developed tensor Fiber CUR decomposition to dramatically lower the computational complexity. The performance advantage of \alg\ is empirically verified against the state-of-the-arts on the synthetic datasets and is further demonstrated on the real-world application such as color video background subtraction.
\end{abstract}

\section{Introduction}
Robust principal component analysis (RPCA) \cite{candes2011robust} is one of the fundamental dimension reduction methods for data science. It tolerances extreme outliers, to which standard PCA is very sensitive. In particular, RPCA aims to reconstruct a low-rank matrix $\BL_\star$ and a sparse outlier matrix $\BS_\star$ from the corrupted observation 
\vspace{-1.5mm}
\begin{equation} \label{eq:rpca}
    \BX=\BL_\star+\BS_\star.
    \vspace{-1.5mm}
\end{equation} 
This RPCA model has been widely studied \cite{netrapalli2014non-convex,bouwmans2018applications,cai2019accelerated,cai2020rapid,cai2021robust} and applied to many applications e.g., 
face modeling \cite{wright2008robust}, 
feature identification \cite{hu2019dstpca}, 
NMR signal recovery \cite{cai2021accelerated},  
and video background subtraction \cite{li2004statistical}. However, RPCA can only handle $2$-mode arrays (i.e., matrices) while real data usually presents in the form of a multi-dimensional array (i.e., tensors). For instance, in the application of video background subtraction \cite{li2004statistical}, a color video is naturally a $4$-mode tensor (height, width, frame, and color).  
To apply RPCA, the original tensor data has to be unfolded to a matrix along certain mode(s), resulting in performance degradation due to the structural information loss at matricization. {In addition, it may not be clear how the rank of the unfolded matrix depends on the rank of the original tensor in some tensor rank settings. } 

To take advantage of the natural tensor structure, 
some earlier studies  have the extended RPCA model \eqref{eq:rpca} to the so-called Robust Tensor PCA (RTPCA) model:
\vspace{-1.5mm}
\begin{equation}
    \cX =\cL_\star+\cS_\star,
    \vspace{-1.5mm}
\end{equation}
where $\cL_{\star}\in\mathbb{R}^{d_1\times\cdots\times  d_n}$ is the underlying low-rank tensor and $\cS_{\star}\in\mathbb{R}^{d_1\times\cdots\times  d_n}$ is the underlying sparse tensor. 
In this model, there is no assumption on the outleirs' magnitudes; they can be arbitrary large as long as $\cS_{\star}$ is sparse. 
Note that there exist various definitions of tensor decompositions that lead to various versions of tensor rank.
In this work, we will focus on the setting using multilinear rank.  
Specifically speaking, we aim to solve the non-convex optimization problem:
\vspace{-1.5mm}
\begin{equation}\label{eq:objective}
    \begin{split}
        \minimize_{\cL,\cS} ~~&\|\cX-\cL-\cS\|_\fro \cr
        \subject &~\cL \textnormal{ is low-multilinear-rank},\cr
         &~\cS \textnormal{ is sparse}. 
    \end{split}
    \vspace{-1.5mm}
\end{equation}
One of the major challenges for solving the multilinear rank based TRPCA problem is the high computational cost for computing the Tucker decomposition. If $\cL_\star$ is rank-$(r_1,\cdots,r_n)$, the existing methods, e.g., \cite{huang2014provable,gu2014robust,sofuoglu2018two,hu2020robust}, have computational complexity at least $\cO(nd^n r)$ flops\footnote{For ease of presentation, we take $d_1=\cdots=d_n=:d$ and $r_1=\cdots=r_n=:r$ when discussing complexities throughout the paper.}---they are thus computationally challenging in large-scale\footnote{In our context, `large-scale' refers to large $n$ and/or large $d$.} problems. Thus, it is urgent to develop a highly efficient TRPCA algorithm for time-intensive applications.

\subsection{Related Work and Contributions}

Due to the various definitions of tensor rank, there are different versions of robust tensor decompositions. For example,  \cite{zhang2014novel} proposes the tensor tubal rank based on the tensor SVD (t-SVD). Based on the tubal rank, the TRPCA is formulated by different convex optimization models with different sparsity patterns  \cite{liu2018improved,lu2019tensor}. This type of RTPCA only partially utilizes the inherent correlation of tensor data. Recently, based on the CP rank \cite{HF1927}, \cite{anandkumar2016tensor} proposed a non-convex iterative algorithm to compute a robust tensor CP decomposition by assuming that the underlying tensor $\cL_\star$ is a low CP rank tensor and $\cS_{\star}$ is a sparse tensor.

In tensor analysis, another popular used tensor rank is multilinear rank \cite{HF1928}. Based on the multilinear rank, there is another version of RTPCA that considers the underlying low rank tensor $\cL_{\star}$ to be a low multilinear rank tensor.  Motivated from the fact that the nuclear norm is the convex envelope of the matrix rank with the unit ball of the spectral norm, the work \cite{goldfarb2014robust} proposed the Higher-order RPCA  via replacing the multilinear rank of the tensor $\cL$ by the convex surrogate $\sum_{i=1}^{n}\|\cL_{(i)}\|_{*}$ and thus the convex optimization problem becomes
\begin{equation}
\begin{split}
    \minimize_{\cL,\cS} ~~&\sum_{i=1}^{n}\|\cL_{(i)}\|_{*}+\lambda \|\cS\|_1\\ 
    \subject &~\cL+\cS=\cX,
\end{split}
\end{equation} 
where $\|\cdot\|_*$ is matrix nuclear norm and $\|\cdot\|_1$ is the entry-wise $\ell_1$ norm.

In this work, we consider the TRPCA problem under the setting of multilinear rank. Our main contributions are two-fold:
\begin{enumerate}
    \item We propose a novel non-convex approach, coined Robust Tensor CUR (\alg), for large-scale TRPCA problems. \alg\ uses an alternating projection framework and employs a novel mode-wise tensor decomposition \cite{cai2021mode} for fast low-rank tensor approximation. The computational complexity of \alg\ is as low as $\cO(n^2dr^2\log^2(d)+n^2r^{n+1}\log^{n+1}(d))$, which is substantially lower than the state-of-the-art.
    
    \item The empirical advantages of \alg\ are verified on both synthetic and real-world datasets. In particular, we show that \alg\ has the best speed performance compared to the state-of-the-art of both tensor and matrix RPCA. 
    Moreover, we also show that tensor methods outperform matrix methods, in terms of reconstruction quality, under certain outlier patterns. 
\end{enumerate}

\subsection{Notation}
We denote tensors, matrices, vectors, and scalars in different typeface for clarity. More specifically, calligraphic capital letters (e.g., $\cX$) are used for tensors,  capital letters (e.g., $X$) are used for matrices, lower boldface letters (e.g., $\mathbf{x}$) for vectors, and regular letters (e.g., $x$) for scalars. We use $X(I,:)$ and $X(:,J)$ to denote the row and column submatrices with indices $I$ and $J$, respectively. $\mathcal{X}(I_1,\cdots,I_n)$ denotes the subtensor of $\mathcal{X}$ with indices $I_k$ at mode $k$. A single element of a tensor is denoted by $\mathcal{X}_{i_1,\cdots, i_n}$. Moreover, $\left\|\mathcal{X}\right\|_{\infty}=\max_{i_1,\cdots,i_n}|\mathcal{X}_{i_1,\cdots,i_n}|$ and $\left\|\mathcal{X}\right\|_\fro=\sqrt{\sum_{i_1,\cdots, i_n}\mathcal{X}_{i_1,\cdots,i_n}^2}$ denote the infinity norm and Frobenius norm, respectively. $X^\dagger$ denotes the Moore-Penrose pseudoinverse.
The set of the first $d$ natural numbers is denoted by $[d]:=\{1,\cdots,d\}$.

\section{Preliminaries}\label{SEC:Prelim}
A tensor is a multidimensional array whose dimension is called the \emph{order} or \emph{mode}. The space of real tensors of order $n$ and sizes $(d_1,\cdots, d_n)$ is denoted as $\mathbb{R}^{d_1\times  \cdots\times d_n}$. 
 We first review some basic tensor properties.
\subsection{Tensor Operations}

\begin{definition}[Tensor Matricization/Unfolding]
An $n$-mode tensor $\mathcal{X}$ can be matricized, or reshaped into a matrix, in $n$ ways by unfolding it along each of the $n$ modes.  
The mode-$k$ matricization/unfolding of tensor $\mathcal{X}\in\mathbb{R}^{d_1\times  \cdots\times d_n}$ is the matrix denoted by
\begin{equation}
\mathcal{X}_{(k)}\in\mathbb{R}^{d_k\times\prod_{j\neq k}d_j}
\end{equation}
whose columns are composed of all the vectors obtained from $\mathcal{X}$ by fixing all indices except for the $k$-th dimension.  The mapping $\mathcal{X}\mapsto \mathcal{X}_{(k)}$ is called the mode-$k$ unfolding operator.
\end{definition}

\begin{definition}[Mode-$k$ Product]~
Let $\mathcal{X}\in\mathbb{R}^{d_1\times \cdots\times d_n}$ and $A\in\mathbb{R}^{J\times d_k}$. The $k$-th mode multiplication between $\mathcal{X}$ and $A$ is denoted by $\mathcal{Y}=\mathcal{X}\times_k A$, with 
\begin{equation}
\begin{split}
&~\mathcal{Y}_{i_1,\cdots,i_{k-1},j,i_{k+1},\cdots,i_{n}}\\
=&~\sum_{s=1}^{d_k}\mathcal{X}_{i_1,\cdots,i_{k-1},s,i_{k+1},\cdots,i_{n}}A_{j,s}.
\end{split}
\end{equation}
Note this can be written as a matrix product by noting that $\mathcal{Y}_{(k)}=A \mathcal{X}_{(k)}$. If we have multiple tensor matrix products from different modes, we  use the notation $\mathcal{X}\times_{i=t}^{s} A_i$ to denote the product $\mathcal{X}\times_{t}A_{t}\times_{t+1}\cdots\times_{s}A_{s}$.
\end{definition}

\begin{definition}[Tucker decomposition and Tucker/Multilinear Rank]
Given an $n$-order tensor $\mathcal{X}$, the tuple $(r_1,\cdots, r_n)\in\mathbb{N}^{n}$ is called the Tucker rank of the tensor $\mathcal{X}$, where $r_k=\text{rank}(\mathcal{X}_{(k)})$ i.e., $r_k$ is the column rank of the unfolding of the tensor $\mathcal{X}$ from mode-$k$.  Its Tucker decomposition is defined as an approximation of a  core tensor $\mathcal{C}\in\mathbb{R}^{R_1\times \cdots \times R_n}$ multiplied by $n$ factor matrices $A_k\in\mathbb{R}^{d_k\times R_k}$ (whose columns are usually orthonormal), $k=1,2,\cdots,n$ along each mode, such that
\begin{equation}\label{eqn:Tucker_D}
\mathcal{X}\approx\mathcal{C}\times_{i=1}^n A_i.
\end{equation}
If  Equation \eqref{eqn:Tucker_D} holds and   $R_k=r_k$ for all $k=1,\cdots,n$, then we call the decomposition an exact Tucker decomposition of $\mathcal{X}$. 
\end{definition}

\begin{remark} \label{remark:hosvd}
  Higher-order singular value decomposition (HOSVD) \cite{de2000best} is  a specific orthogonal Tucker decomposition which is popularly used in the literature.
\end{remark}

\subsection{Tensor Fiber CUR Decompostions}
\begin{figure}[t]
    \centering
    \includegraphics[width=.8\linewidth]{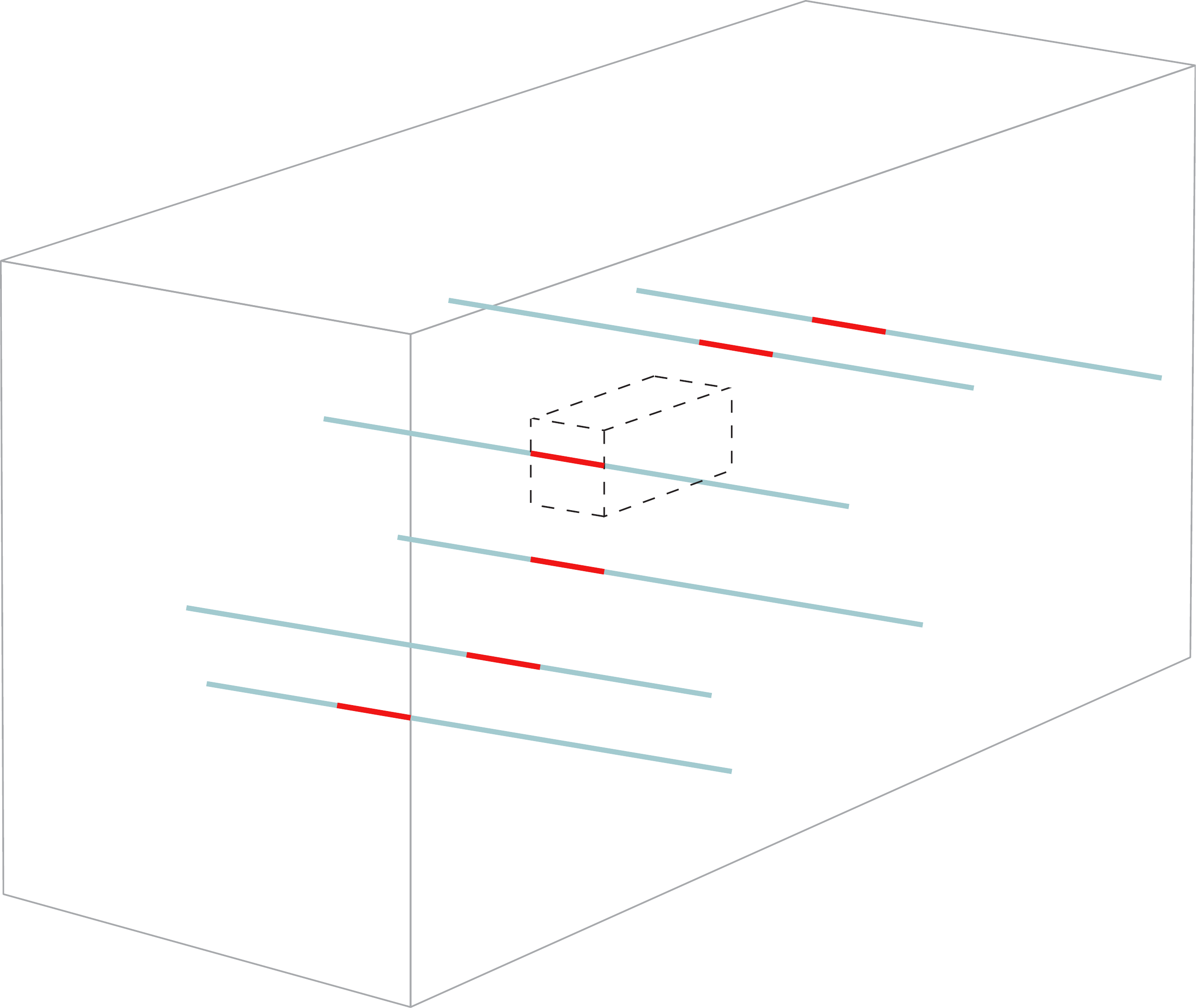}
    \caption{(\cite[Figure 2]{cai2021mode}). Illustration of the Fiber CUR Decomposition of Theorem~\ref{thm: CUR Char} in which $J_i$ is not necessarily related to $I_i$.  The lines correspond to rows of $C_2$, and red indices within  correspond to rows of $U_2$.  Note that the lines may (but do not have to) pass through the core subtensor $\cR$ outlined by dotted lines.  Fibers used to form $C_1$ and $C_3$ are not shown for clarity.}
    \label{FIG:TensorCURIndependent}
\end{figure}

CUR decompositions for matrices have been actively studied  \cite{Goreinov,HH2020}. 
For a matrix $X\in\mathbb{R}^{d_1\times d_2}$, let $C$ be  a column submatrix of $X$ with column indices $J$ , $R$ be a row submatrix of $X$ with row indices $I$, and $U=X(I,J)$. The theory of CUR decompositions states that $X=CU^\dagger R$ if $\rank(U)=\rank(X)$. 
The first extension of CUR decompositions to tensors  involved a single-mode unfolding of 3-mode tensors \cite{MMD2008}. 
Later, \cite{caiafa2010generalizing} proposes a different variant of tensor CUR that accounts for all modes. Recently, \cite{cai2021mode} dubs these decompositions with more descriptive monikers, namely Fiber and Chidori CUR decompositions. Later in this paper, we will employ the tensor Fiber CUR decomposition (see Figure~\ref{FIG:TensorCURIndependent} for illustration) to accelerate a key step in the proposed algorithm. 
For the reader's convenience, we state the characterization of the Fiber CUR decomposition below.
\begin{theorem}[{\cite[Theorem~3.3]{cai2021mode}}]\label{thm: CUR Char}
Let $\mathcal{A}\in\mathbb{R}^{d_1\times\cdots\times d_n}$ with multilinear rank $(r_1,\dots,r_n)$. Let $I_i\subseteq [d_i]$ and $J_i\subseteq[\prod_{j\neq i}d_j]$. Set $\mathcal{R}=\mathcal{A}(I_1,\cdots,I_n)$, $C_i=\mathcal{A}_{(i)}(:,J_i)$ and $U_i=C_i(I_i,:)$. 
Then the following statements are equivalent:
\begin{enumerate}[label=(\roman*)]
    \item  \label{CUR Char:item1} $\rank(U_i)=r_i$,
    \item  \label{CUR Char:item2} $\mathcal{A}=\mathcal{R}\times_{i=1}^{n}(C_iU_i^\dagger)$,
    \item  \label{CUR Char:item3} $\rank(C_i)=r_i$ for all $i$ and the multilinear rank of $\mathcal{R}$ is $(r_1,\cdots,r_n)$.
\end{enumerate}
In particular, \ref{CUR Char:item2} is called tensor Fiber CUR decomposition.
\end{theorem}
In addition, according to \cite[Corollary 5.2]{hamm2020stability},  if one uniformly samples indices $I_i$ and $J_i$ with size $|I_i|= \cO(r_i\log(d_i))$ and $|J_i|=\cO\left(r_i\log(\prod_{j\neq i}d_j)\right)$,
then  $\rank(U_i)=r_i$ holds for all $i$ with high probability under some mild assumptions. Thus, the tensor Fiber CUR decomposition holds and its computational complexity is dominated by computing the pseudoinverse of $U_i$. Given the dimension of $U_i$, computing the pseudoinverse costs $\cO\left(r^2(n-1)\log^2(d)\right)$ flops, thus tensor Fiber CUR decomposition costs $\cO\left(r^2n^2\log^2(d)\right)$ flops. In contrast, HOSVD costs $\cO(rd^n)$ flops.

\section{Proposed Approach}
In this section, we propose a fast approach, called Robust Tensor CUR (\alg), for non-convex TRPCA problem \eqref{eq:objective}. \alg\ is developed in a framework of alternating projections: (I) First,  we project $\cX-\cL^{(k)}$ onto the space of sparse tensors to update the estimate of outliers (i.e., $\cS^{(k+1)}$); (II) then we project the less corrupted data $\cX-\cS^{(k+1)}$ onto the space of low-multilinear-rank tensors to update the estimate (i.e., $\cL^{(k+1)}$). The key for our algorithm acceleration is using tensor Fiber CUR decomposition for inexact low-multilinear-rank tensor approximation in Step (II), which provides much better computational complexity than the standard HOSVD. 
Consequently, in Step~(I), this inexact approximation allows us to estimate only the outliers that lie in the smaller subtensors and fibers that Fiber CUR decomposition samples. 
\alg\ is summarized in Algorithm~\ref{alg:TRCUR}. 
Now, we will discuss the details of our approach and start with Step~(II).

\begin{algorithm}[t]
\caption{\textbf{R}obust \textbf{T}ensor  \textbf{CUR} (\alg)}
 \label{alg:TRCUR}
\begin{algorithmic}[1]
\State \textbf{Input: }{$\cX=\cL_\star+\cS_\star\in\mathbb{R}^{d_1\times \cdots \times d_n}$: observed tensor; $(r_1, \cdots, r_n)$: underlying multilinear rank of $\cL_\star$; $\varepsilon$: targeted precision; $\zeta^{(0)},\gamma$: thresholding parameters; $\{|I_i|\}_{i=1}^n,\{|J_i|\}_{i=1}^n$: cardinalities for sample indices. }
\State \textbf{Initialization:} $\cL^{(0)} = \mathbf{0},\cS^{(0)} = \mathbf{0},k=0$ 
\State Uniformly sample the indices $\{I_i\}_{i=1}^n, \{J_i\}_{i=1}^n$ 
\While {$e^{(k)} > \varepsilon$} \qquad\qquad\quad {\color{OliveGreen}// $e^{(k)}$ is defined in \eqref{eq:rel_err}}
\State 
 {\color{OliveGreen} (Optional)} Resample the indices $\{I_i\}_{i=1}^{n}, \{J_i\}_{i=1}^{n}$ 
\State {\color{OliveGreen} // Step (I): Updating $\cS$} 
  \State $\zeta^{(k+1)} = \gamma\cdot \zeta^{(k)}$ 
  \State $\cS^{(k+1)} = \mathrm{HT}_{\zeta^{(k+1)}}(\tens{X}-\tens{L}^{(k)})$  

\State {\color{OliveGreen} // Step (II): Updating $\cL$} 
\State $\cR^{(k+1)}= (\cX-\cS^{(k+1)})(I_1,\cdots,I_n)$ 
 \For{$i = 1,\cdots,n$}
 \State $C_i^{(k+1)} = (\tens{X}-\cS^{(k+1)})_{(i)}(:,J_i)$ 
 \State $U_i^{(k+1)} = \mathrm{SVD}_{r_i}(C_i^{(k+1)}(I_i,:))$ 
\EndFor
 \State $\cL^{(k+1)} = \cR^{(k+1)}\times_{i=1}^{n} C_i^{(k+1)}\left(U_{i}^{(k+1)}\right)^{\dagger}$ 
 \State $k = k+1$ 
 \EndWhile
 \State \textbf{Output: }{$\cR^{(k)}, C_i^{(k)},U_i^{(k)}$ for $i=1,\cdots,n$: the estimates of  the tensor Fiber CUR decomposition of $\cL_{\star}$.
 }
 
\end{algorithmic}
\end{algorithm}

\subsection{Updating $\cL$} 
SVD is the most standard method for low-rank approximation in the matrix setting. Similarly, in the literature, HOSVD is the standard method for low-multilinear-rank approximation under our multilinear rank setting. However, HOSVD is computationally  expansive when the problem scale is large. Inspired by the recent development on tensor CUR decomposition \cite{cai2021mode}, we employ Fiber CUR decomposition for accelerated inexact low-multilinear-rank tensor approximations and update the estimate by setting 
\begin{equation} \label{eq:L=CUR}
\cL^{(k+1)} = \cR^{(k+1)}\times_{i=1}^{n} C_i^{(k+1)}\left(U_{i}^{(k+1)}\right)^{\dagger},
\end{equation}
where 
\begin{equation} \label{eq:R_and_C_and_U}
\begin{split}
    \cR^{(k+1)} &= (\cX-\cS^{(k+1)})(I_1,\cdots,I_n), \cr
    C_i^{(k+1)} &= (\cX-\cS^{(k+1)})_{(i)}(:,J_i), \cr
    U_i^{(k+1)} &= \mathrm{SVD}_{r_i}(C_i^{(k+1)}(I_i,:)),
\end{split}
\end{equation}
for all $i$. Therein, $\mathrm{SVD}_{r_i}$ denotes the truncated rank-$r_i$ matrix singular value decomposition {that is used to enforce the low-multilinear-rank constraint.} Recall the aforementioned complexity for computing Fiber CUR decomposition is $\cO\left(r^2n^2\log^2(d)\right)$. Note that there is no need to compute the full $\cX-\cS^{(k+1)}$ but merely the sampled subtensors. 
Similarly, forming the entire $\cL$ is never needed through \alg---we only need to keep the decomposed tensor components and form the sampled subtensors of $\cL$ accordingly (see Section \ref{sec:cc} for details). That is, \eqref{eq:L=CUR} is provided for  explanation but should never be fully computed in an efficient implementation of \alg.

\subsection{Updating $\cS$} 
We consider the simple yet effective hard thresholding operator $\mathrm{HT}_\zeta$ for outlier estimation, which is defined as:
\begin{equation}
    (\mathrm{HT}_{\zeta}\cX)_{i_1,\cdots,i_n} =
    \begin{cases}
    \cX_{i_1,\cdots,i_n}, & \quad|\cX_{i_1,\cdots,i_n}| >\zeta;\\
    0,  & \quad\mbox{otherwise.}
    \end{cases}
\end{equation}
As shown in \cite{cai2019accelerated,cai2020rapid,netrapalli2014non-convex}, with a properly chosen thresholding value, $\mathrm{HT}_{\zeta}$ is effectively a projection operator onto the support of $\cS_\star$. More specifically, we update
\begin{equation} \label{eq:S=HT(X-L)}
  \tens{S}^{(k+1)} = \mathrm{HT}_{\zeta^{(k+1)}}(\cX-\cL^{(k)}).
\end{equation}
If $\zeta^{(k+1)}=\|\cL_\star-\cL^{(k+1)}\|_\infty$ is chosen, then we have $\supp(\cS^{(k+1)})\subseteq\supp(\cS_\star)$ and $\|\cS_\star-\cS^{(k+1)}\|_\infty\leq 2\|\cL_\star-\cL^{(k+1)}\|_\infty$. Empirically, we find that iteratively decaying thresholding values
\begin{equation}
    \zeta^{(k+1)} = \gamma \cdot \zeta^{(k)}
\end{equation}
provide superb performance with carefully tuned $\gamma$ and $\zeta^{(0)}$. Note that a favorable choice of $\zeta^{(0)}$ is $\|\cL_\star\|_\infty$, which can be easily accessed/estimated in many applications. The decay factor $\gamma\in(0,1)$ should be tuned according to the difficulty of the TRPCA problem, e.g., those problems with higher rank, more dense outliers, or large condition number are harder. For successful reconstruction, the harder problems require larger $\zeta$ and \textit{vice versa}. We observe that $\gamma\in[0.6,0.9]$ generally performs well.

\subsection{Computational Complexity} \label{sec:cc}
As mentioned, the complexity for computing a Fiber CUR decomposition is very low, thus the dominating steps in \alg\ are the hard thresholding operator and/or the tensor/matrix multiplications. 
We again remark that only the sampled subtensors and fibers are needed when computing \eqref{eq:R_and_C_and_U}. Thus, we merely need to estimate the outliers on these subtensors and fibers, and \eqref{eq:S=HT(X-L)} should not be fully executed. Instead, we only compute
\begin{equation} \label{eq:partial S update}
\begin{split}
    \cS^{(k+1)}(I_1,\cdots,I_n) &= \mathrm{HT}_{\zeta^{(k+1)}}((\cX-\cL^{(k)})(I_1,\cdots,I_n)), \cr
    \cS^{(k+1)}_{(i)}(:,J_i) &= \mathrm{HT}_{\zeta^{(k+1)}}((\cX-\cL^{(k)})_{(i)}(:,J_i)). 
\end{split}
\end{equation}
for all $i$, which leads to a total complexity of $\cO(r^{n}\log^n(d)+n^2dr\log(d))$ in this step.
Not only can we save the computational complexity on hard thresholding but also much smaller subtensors of $\cL^{(k)}$ need to be formed in  \eqref{eq:partial S update}. We can form the required subtensors from the stored Fiber CUR components, which is much cheaper than forming and saving the whole $\cL^{(k)}$. 
In particular, the total complexity for forming $\cL^{(k)}(I_1,\cdots,I_n)$ and $\{\cL^{(k)}_{(i)}(:,J_i)\}_{i=1}^n$ is $\cO(n^2dr^2\log^2(d)+n^2r^{n+1}\log^{n+1}(d))$ and it dominates the overall complexity of \alg. 

Moreover, for time saving purposes, we may avoid {computing} the Frobenius norm of the full tensor when computing the relative error for the stopping criterion. In \alg, we adjust the relative error formula to be  
\begin{equation} \label{eq:rel_err}
    e^{(k)}=  \frac{\|\cE^{(k)}(I_1,\cdots,I_n)\|_\fro+\sum_{i=1}^n\|\cE^{(k)}_{(i)}(:,J_i)\|_\fro}{\|\cX(I_1,\cdots,I_n)\|_\fro+\sum_{i=1}^n\|\cX_{(i)}(:,J_i)\|_\fro}
\end{equation}
where $\cE^{(k)}=\cX-\cL^{(k)}-\cS^{(k)}$, 
so that it does not use any extra subtensor or fiber but only those we already formed.

\subsection{Two Variants of \alg} We consider two variants of \alg, namely \algf\ and \algr. \algf\ uses fixed sample indices through all iterations while \algr\ resamples $\{I_i\}_{i=1}^n$ and $\{J_i\}_{i=1}^n$ in every iteration. \algf\ requires minimal data accessibility and runs slightly faster. \algr\ accesses more data and takes some extra computing; for example, the denominator of \eqref{eq:rel_err} has to be recomputed per iteration. On the other hand, accessing more redundant data means \algr\ has a better chance to correct any ``unlucky" sampling over the iterations, thus we expect that \algr\ has superior outlier tolerance than \algf.
We recommend \algf\ if the data accessibility or computation ability is limited. Otherwise, if better outlier tolerance is desired and data is easily accessible, then \algr\ is recommended.

\section{Numerical Experiments}

\begin{figure}[t]
    \centering
	\includegraphics[width=0.49\linewidth]{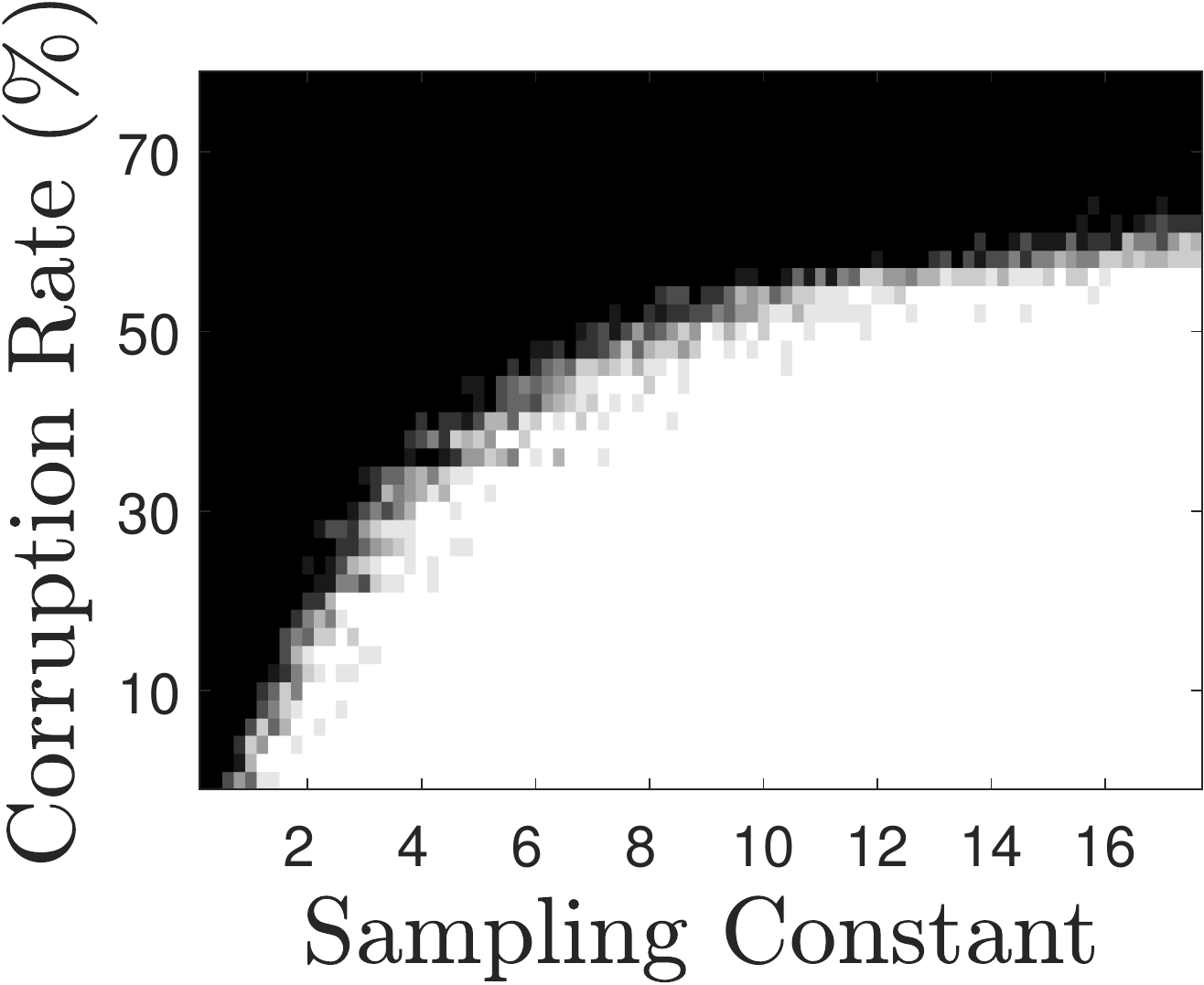}
 \hfill
 		\includegraphics[width=0.49\linewidth]{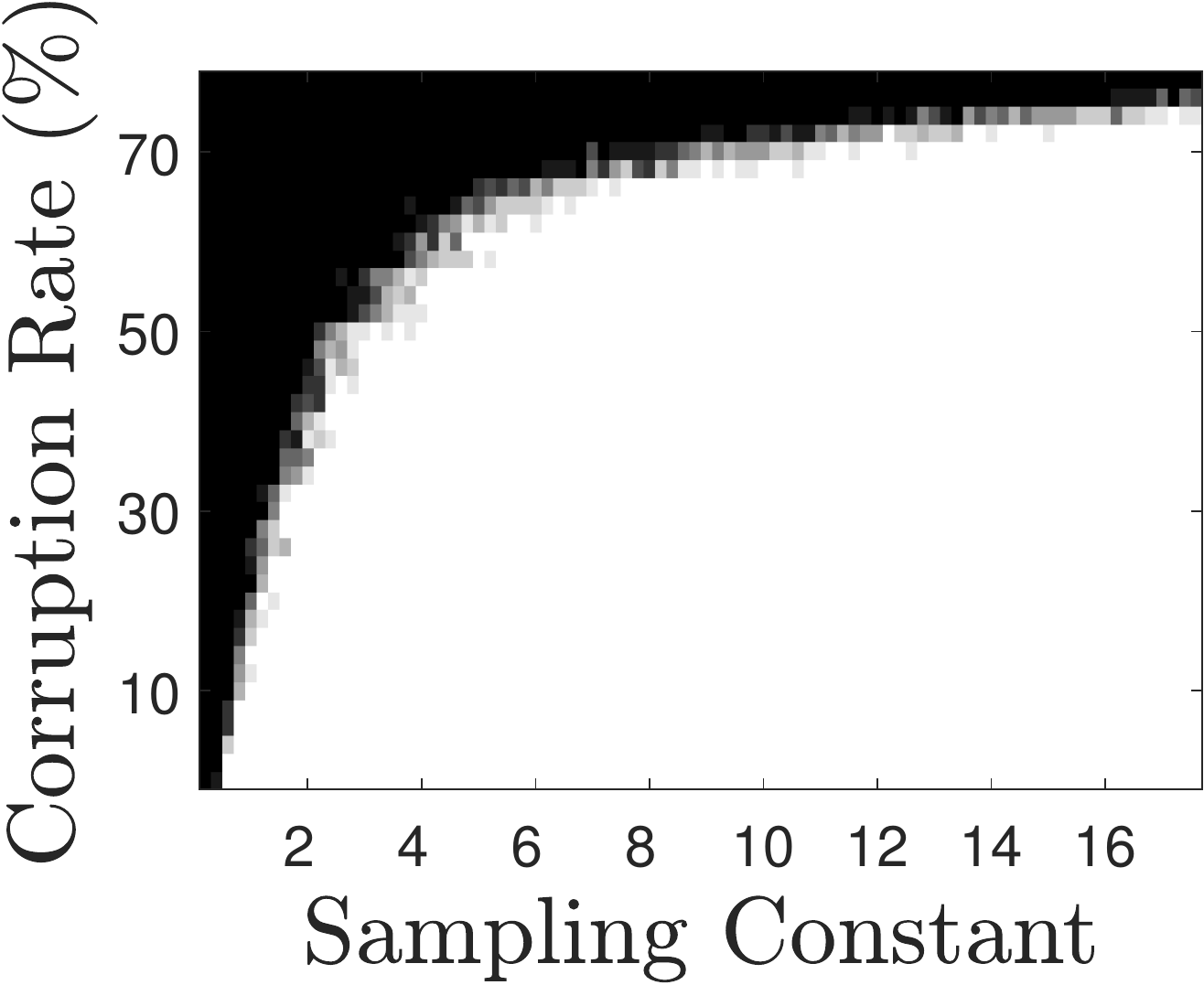}\\
    \includegraphics[width=0.49\linewidth]{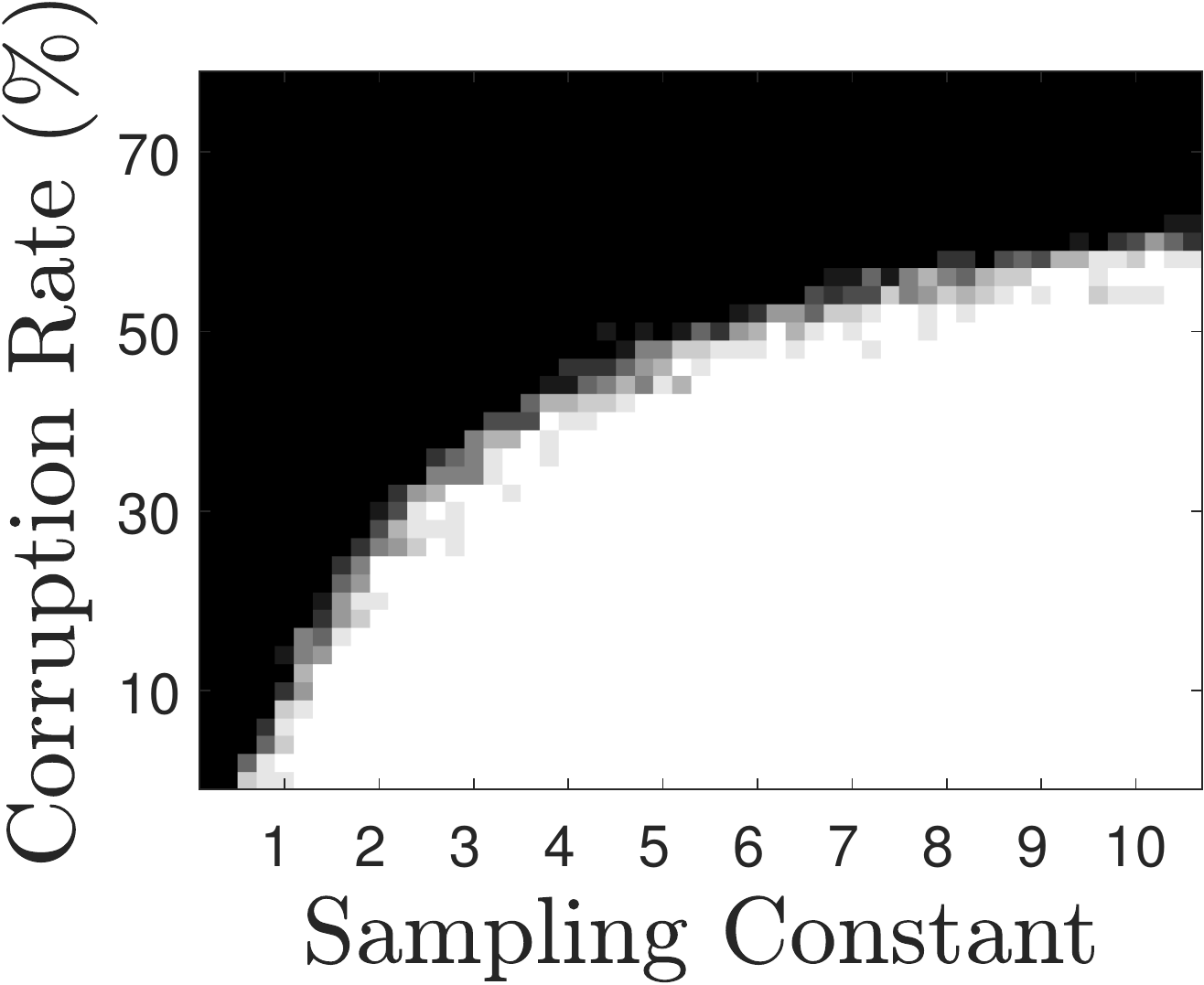}
\hfill
\includegraphics[width=0.49\linewidth]{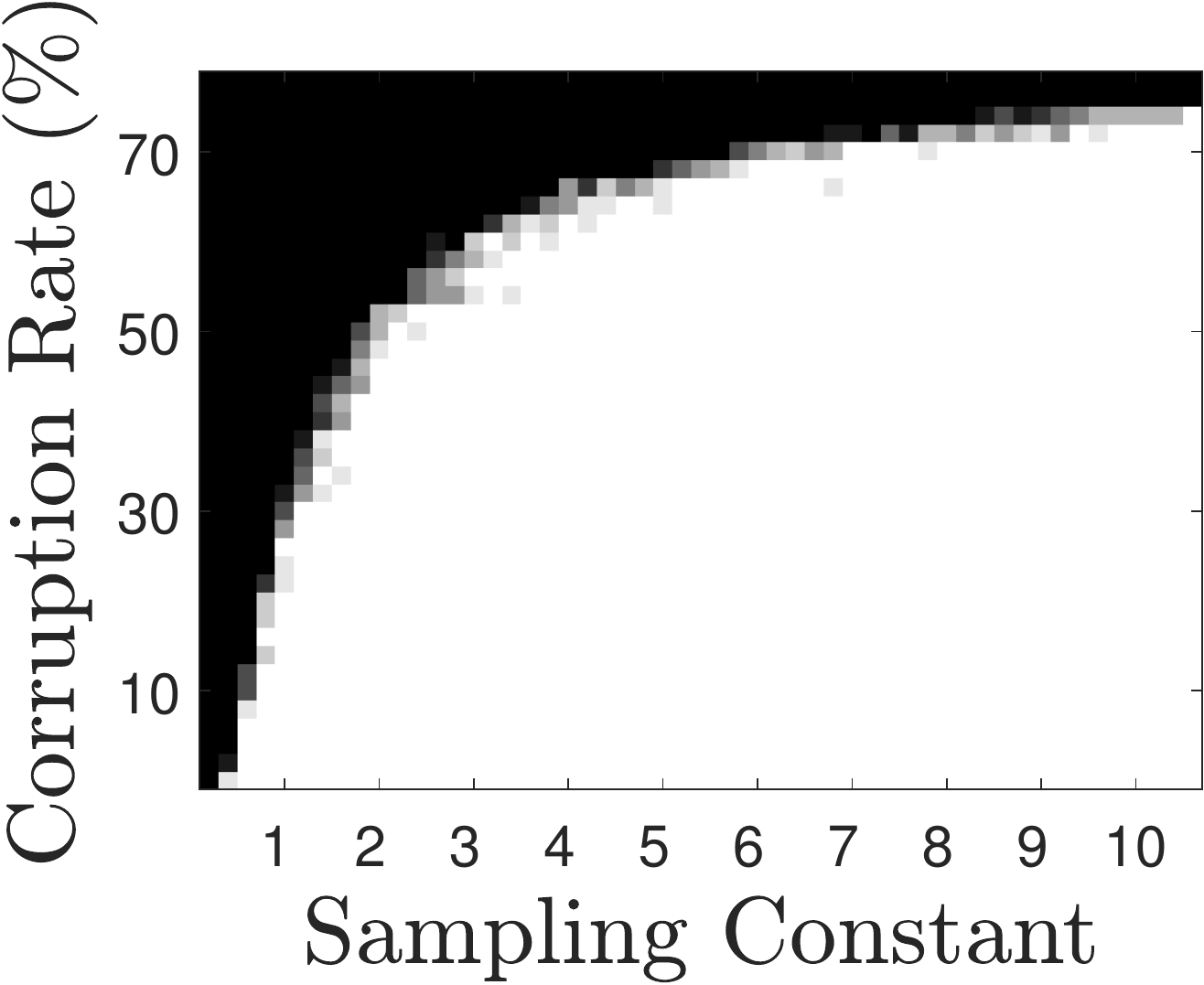}\\
	\includegraphics[width=0.49\linewidth]{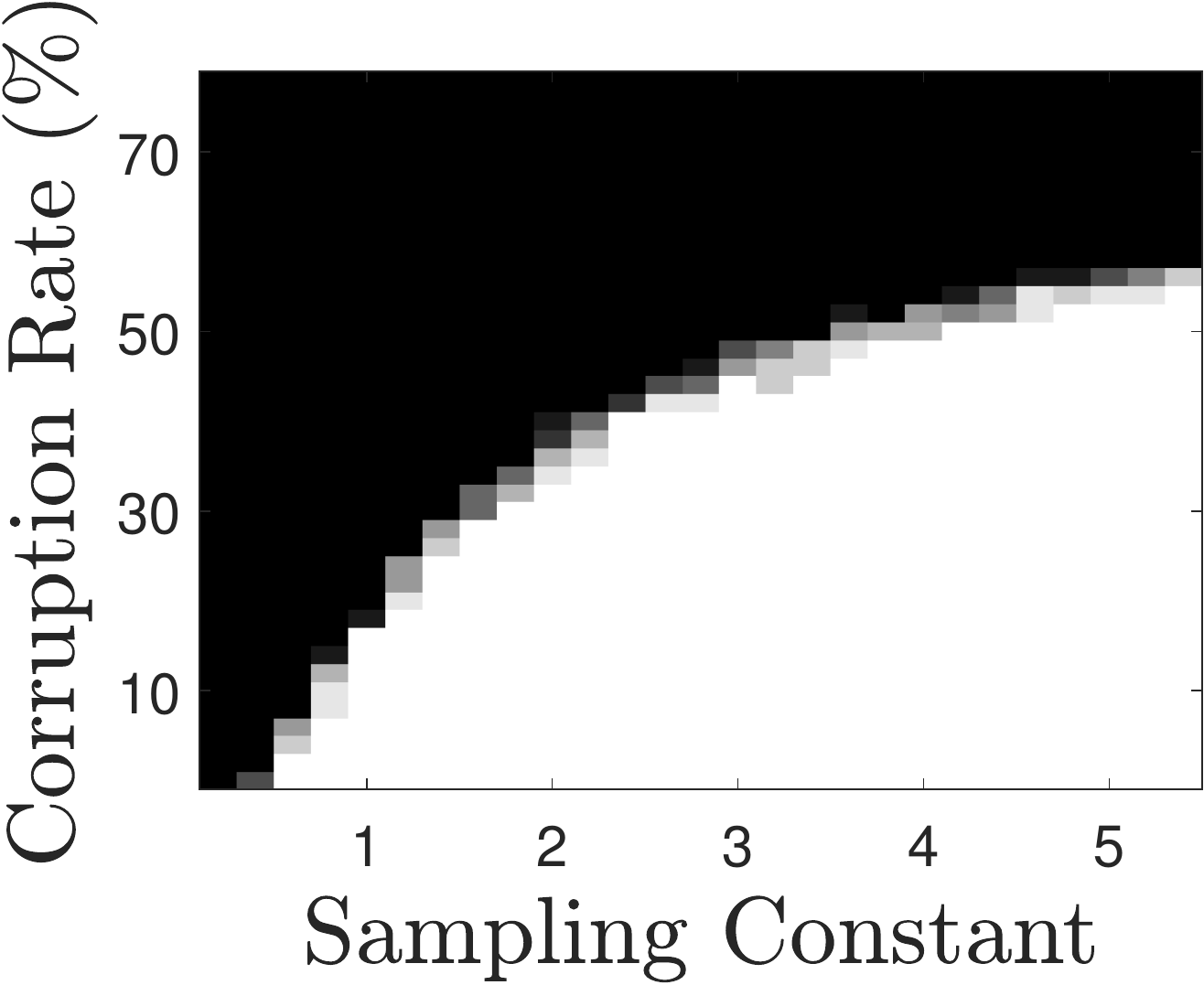}
	\hfill
	\includegraphics[width=0.49\linewidth]{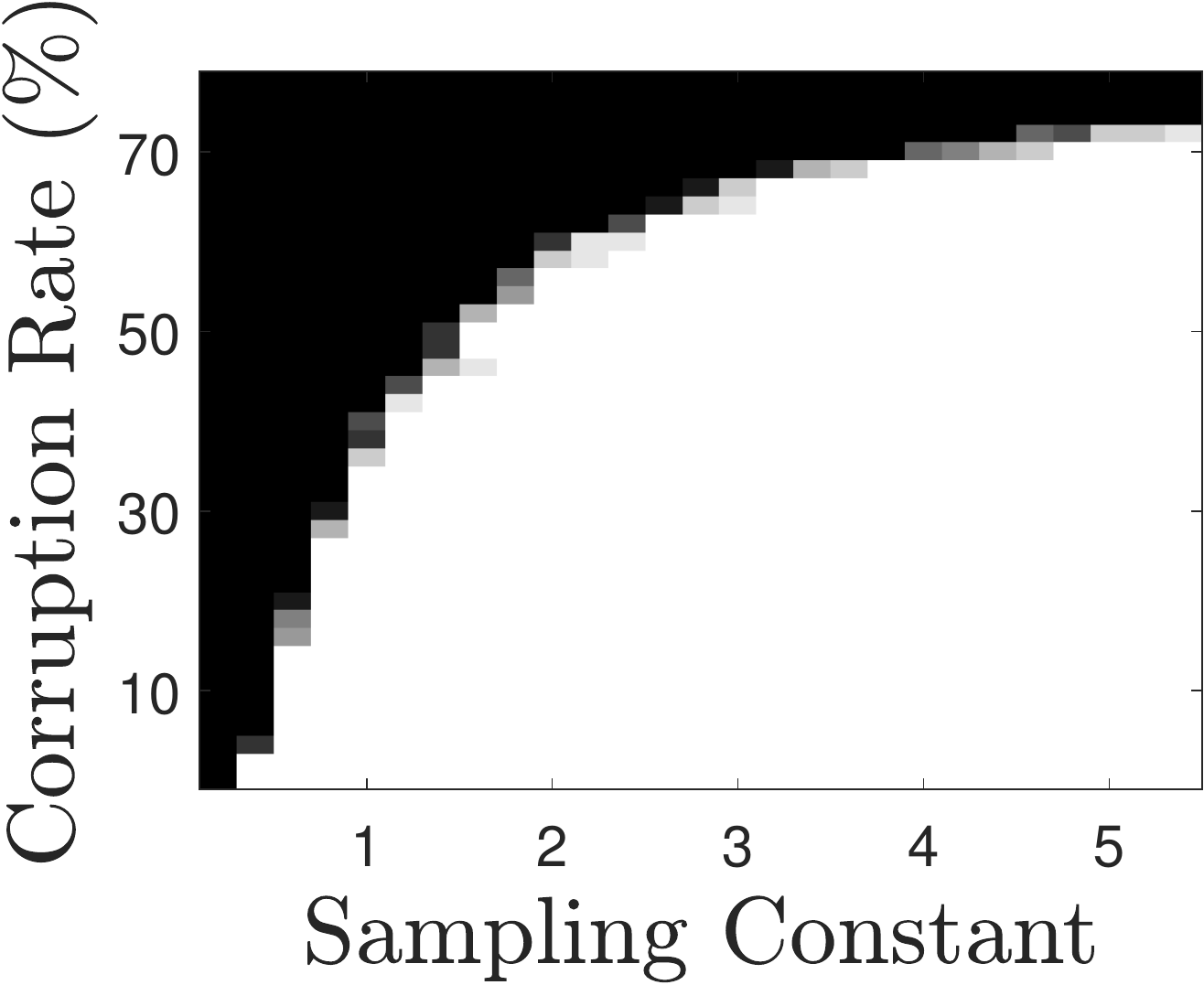}
	 
	  \caption{Empirical phase transition in corruption rate $\alpha$ and sampling constant $\upsilon$. \textbf{Left}: \algf. \textbf{Right}: \algr. \textbf{Top}:
$r = 3$. \textbf{Middle}: $r = 5$. \textbf{Bottom}: $r = 10$.} \label{FIG:phase}
\end{figure}

In this section, we conduct numerical experiments to verify the empirical performance of \alg\ against the state-of-the-art robust matrix/tensor PCA algorithms:  Alternating Direction Method of Multipliers for TRPCA (ADMM) \cite{lu2019tensor}, Accelerated Alternating Projections for RPCA (AAP) \cite{cai2019accelerated}, and Iterative Robust CUR for RPCA (IRCUR) \cite{cai2020rapid}. 
Both proposed variants, \algf\ and \algr, will be evaluated. 
Section~\ref{sec:synthetic} contains two synthetic experiments: 
(i) We study the empirical relation between the outlier tolerance and sample size for \alg. (ii) We show the speed advantage of \alg\ compared to the state-of-the-art methods. 
In Sections~\ref{sec:video}, 
we successfully apply \alg\ to two real-world problems, color video background subtraction and face modeling. 

We utilize the codes of all compared algorithms from the authors' websites, and the parameters are hand tuned for their best performance. For \alg, we sample $|I_i|=\upsilon r_i\log(d_i)$ and $|J_i|=\upsilon r_i \log(\prod_{j\neq i}d_j)$ for all $i$, and $\upsilon$ is called the \textit{sampling constant} through this section. 
All the tests are executed from Matlab R2020a on an Ubuntu workstation with Intel i9-9940X CPU and 128GB RAM.

\begin{figure}[t]
    \centering
    \subfloat{\includegraphics[width = 0.8\linewidth]{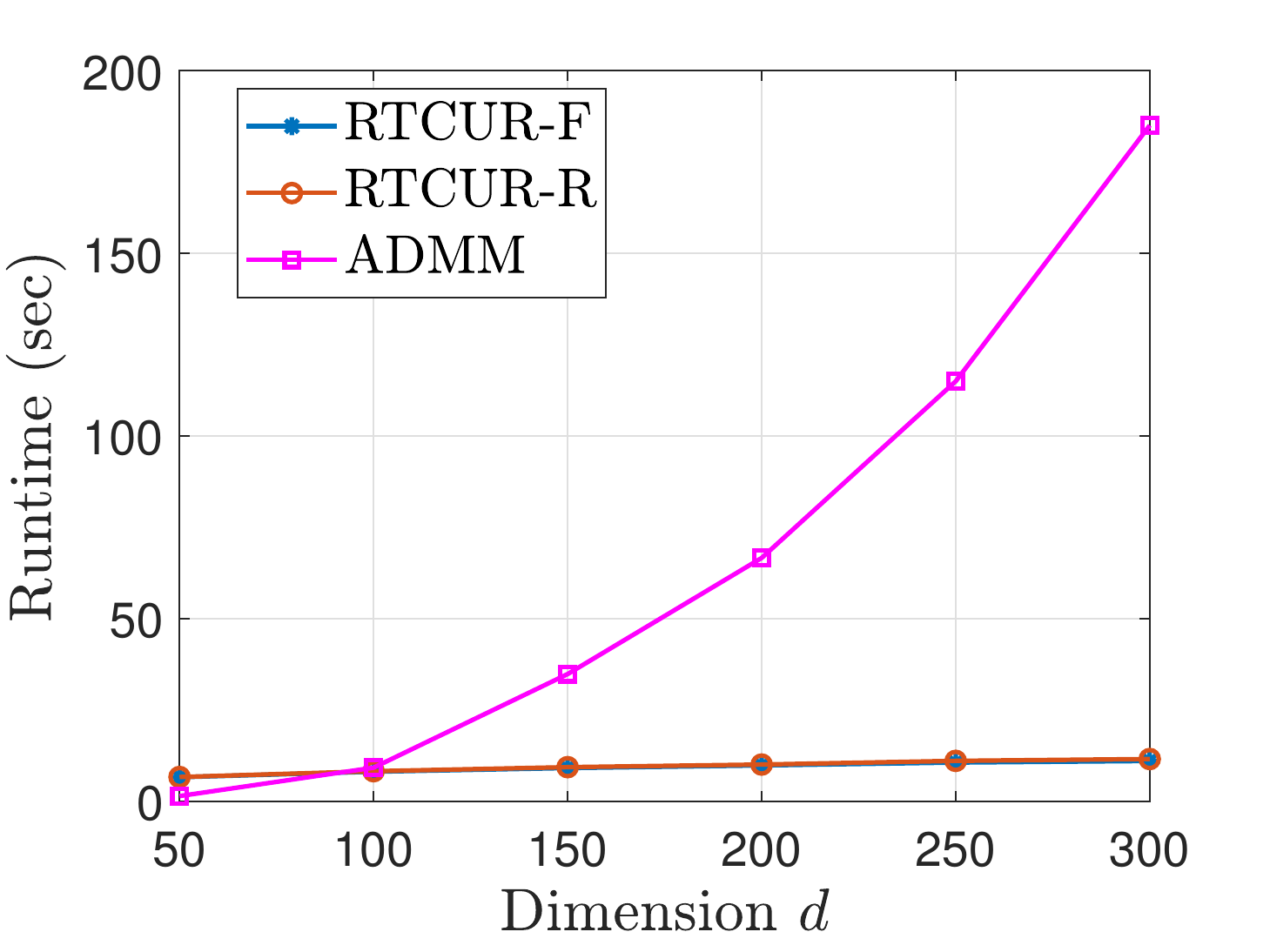}}
    \\
    \subfloat{\includegraphics[width = 0.8\linewidth]{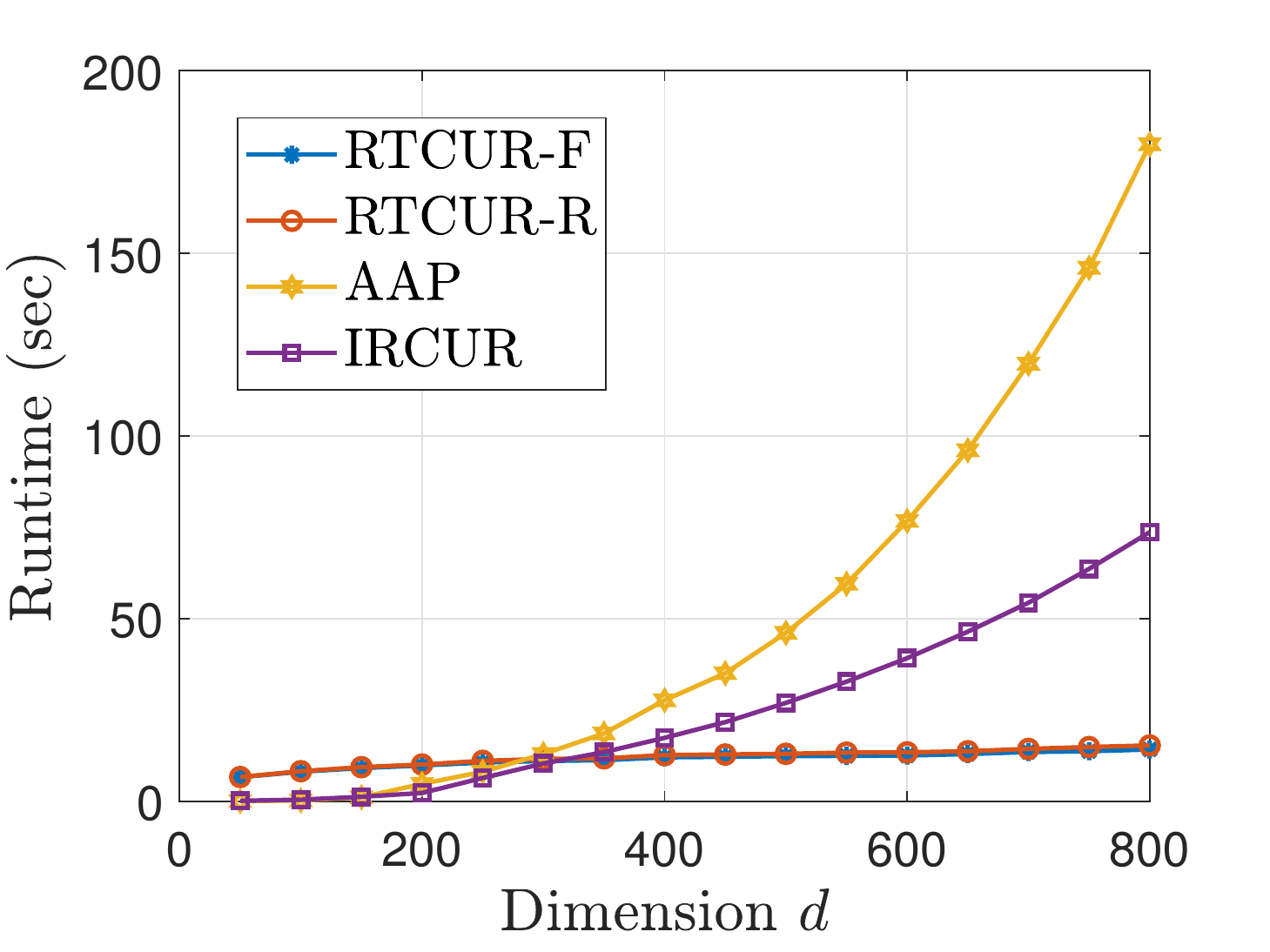}}
    \caption{Runtime vs.~dimension comparison among \algf, \algr, ADMM, AAP,  and IRCUR. Multilinear rank $(3,3,3)$. The ADMM method proceed relatively slow for larger tensor so we only test the ADMM runtime for tensor size smaller than 300.}
    \label{fig:CUR_time}
\end{figure}

\begin{table*}[t]
\caption{Video information and runtime comparison for color video background subtraction task.
}
\vspace{-0.1cm}
\label{table:video}
 \centering
 \begin{tabular}{ |c||c|c|c|c|c|c|} 
\hline
          &video &   \multicolumn{5}{c|}{runtime (sec)} \cr
 \cline{3-7}

~             & size                                                   & {RTCUR-F} &{RTCUR-R}  &{ADMM \cite{lu2019tensor}} &{AAP \cite{cai2019accelerated}} & {IRCUR \cite{cai2020rapid}}               \cr
 \hhline {|=||=|=|=|=|=|=|}
\textit{Shoppingmall}
&
$256\times 320\times 3\times 1250$ &\textbf{3.53} & 5.83 & 783.67 & 50.38 & 15.71   \cr
\hline
\textit{Highway}
 &
$240\times 320\times3\times440$  & \textbf{3.15} & 5.47 & 168.55  & 18.10 & 3.87 \cr
\hline
\textit{Crossroad}
 &
$350\times640\times3\times900$  & \textbf{6.15} & 13.33  & 1099.3 & 97.85 & 35.47 \cr
\hline
\textit{Port}
 &
$480\times640\times3\times1000$  & \textbf{7.22} & 8.10  & 3502.8 & 121.2 & 11.98 \cr
\hline
\textit{Parking lot}
 &
$360\times640\times3\times400$  & \textbf{6.30} & 8.91  & 1001.5 & 43.12 & 9.02 \cr
\hline

\end{tabular}
\end{table*}

\subsection{Synthetic Examples} \label{sec:synthetic}

For the synthetic experiments, we use $d:=d_1=\cdots=d_n$ and $r:=r_1=\cdots=r_n$. The observed tensor $\cX=\cL_\star+\cS_\star$. To generate $n$-mode $\cL_\star\in\mathbb{R}^{d\times\cdots\times d}$ with multilinear rank $(r,\cdots,r)$, we take $\cL_\star=\mathcal{Y}\times_1 Y_1 \times_2 \cdots \times_n Y_n$ where $\mathcal{Y}\in\mathbb{R}^{r\times\cdots\times r}$ and $\{Y_i\in\mathbb{R}^{d\times r}\}_{i=1}^n$ are Gaussian random tensor and matrices with standard normal entries. To generate the sparse outlier tensor $\cS_\star$, we uniformly sample $\alpha$ percent entries to be the support of $\cS_\star$ and the values of the non-zero entries are uniformly sampled from the interval $[-\mathbb{E}(|\cL_{i_1,\cdots,i_n}|),\mathbb{E}(|\cL_{i_1,\cdots,i_n}|)]$.

\vspace{0.15in}
\noindent\textbf{Phase transition.}
We study the empirical relation between the outlier corruption rate $\alpha$ and sampling constant $\upsilon$ for \algf\ and \algr. The tests are conducted on $300\times300\times300$ (i.e., $n=3$ and $d=300$) problems with $r=3,5,$ or $10$. For both the variants, the thresholding parameters are set to be $\zeta^{(0)}=\|\cL\|_\infty$ and $\gamma=0.7$. The stopping condition is $e^{{k}}<10^{-5}$ and an example is considered successfully solved if $\|\cL_\star-\cL^{(k)}\|_\fro/\|\cL_\star\|_\fro\leq 10^{-3}$. For each pair of $\alpha$ and $\upsilon$, we generate $10$ test examples. 

The experimental results are summarized in Figure~\ref{FIG:phase}, where a white pixel means all $10$ test examples are successfully solved under the corresponding problem parameter setting and a black pixel means all $10$ test cases fail. 
As expected, one can see \algr\ has slightly better performance than \algf\ and smaller $r$ tolerates more outliers since a larger $r$ leads to a harder problem. Moreover, taking larger $\upsilon$ clearly provides better outlier tolerance. However, larger $\upsilon$ means \alg\ needs to sample larger subtensors and more fibers, which of course leads to more computational time. According to this experiment, we recommend $\upsilon\in [3,5]$ to balance between outlier tolerance and speed.

\vspace{0.15in}
\noindent\textbf{Computational efficiency.}
In this section, we compare the computational efficiency between \alg\ and the state-of-the-art tensor/matrix RPCA algorithms. For matrix methods, we first unfold a $n$-mode $d\times \cdots \times d$ tensor to a $d\times d^{n-1}$ matrix, then solve the matrix RPCA problem with rank $r$. For all tests, we add $10\%$ corruption and set parameters $\upsilon=3$, $\zeta^{(0)}=\|\cL\|_\infty$, $\gamma = 0.7$ for both \algf\ and \algr. The reported runtime is averaged over $10$ trials.

In Figure~\ref{fig:CUR_time}, we consider the problem of $3$-mode TRPCA with varying dimension $d$ and compare the total runtime (all methods halt when $e^{(k)}<10^{-5}$). One can see both variants of \alg\ are substantially faster than the compared algorithms when $d$ is large.

\subsection{Color Video Background Subtraction} \label{sec:video}

We apply \alg\ on the task of color video background subtraction. We consider two color video datasets with static backgrounds: \textit{Shoppingmall} \cite{l}, \textit{Highway} \cite{bouwmans2017scene}, \textit{Crossroad} \cite{bouwmans2017scene}, \textit{Port} \cite{bouwmans2017scene}, and \textit{Parking Lot} \cite{oh2011large}. 
Since a monochromatic frame usually does not have low rank structure \cite{chao2020tensor}, we vectorize each color channel of each frame into a vector and construct a (height $\times$ width)$\times 3 \times$ frames) tensor. The targeted multilinear rank is $\mathbf{r} = (3,3,3)$ for both videos. For those matrix algorithms, we unfold the tensor to a (height $\times$ width)$\times$($3\times$ frames) matrix and rank $3$ is used. We set \alg\ parameters $\upsilon=2$, $\zeta^{(0)}=255$, $\gamma=0.7$ in this experiment.

\begin{figure*}[!ht]
\centering
\vspace{-0.17in}
\subfloat[Original]{\includegraphics[width=0.16\linewidth]{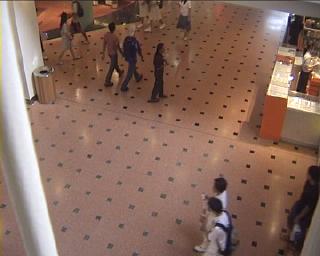}}\hfill
\subfloat[RTCUR-F]{\includegraphics[width=0.16\linewidth]{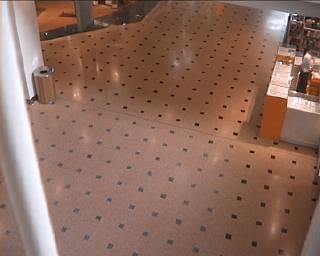}}\hfill
\subfloat[RTCUR-R]{\includegraphics[width=0.16\linewidth]{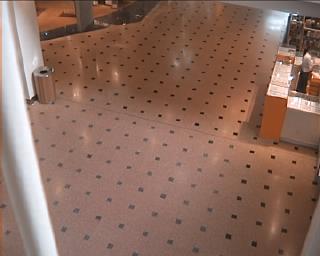}}\hfill
\subfloat[ADMM]{\includegraphics[width=0.16\linewidth]{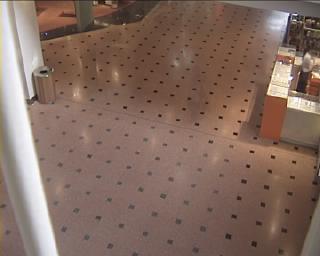}}\hfill
\subfloat[AAP]{\includegraphics[width=0.16\linewidth]{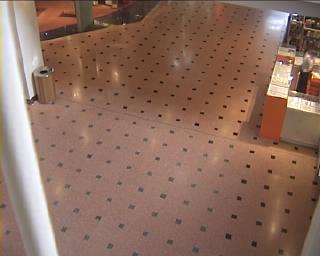}}\hfill
\subfloat[IRCUR]{\includegraphics[width=0.16\linewidth]{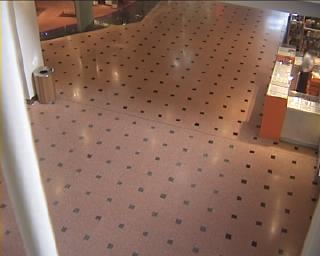}}
\vspace{-0.12in}
\\
\subfloat{\includegraphics[width=0.16\linewidth]{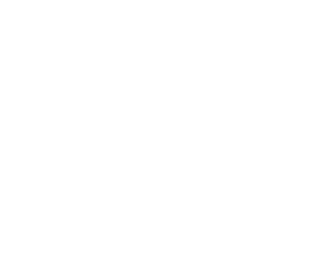}}\hfill
\subfloat{\includegraphics[width=0.16\linewidth]{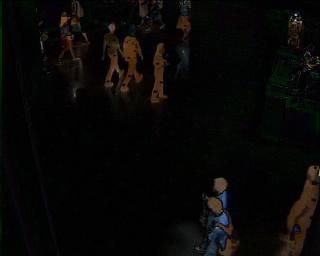}}\hfill
\subfloat{\includegraphics[width=0.16\linewidth]{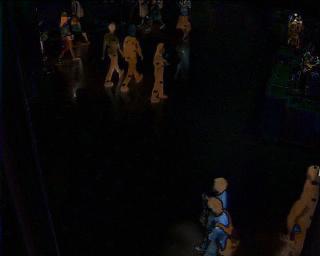}}\hfill
\subfloat{\includegraphics[width=0.16\linewidth]{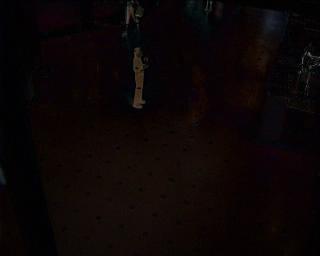}}\hfill
\subfloat{\includegraphics[width=0.16\linewidth]{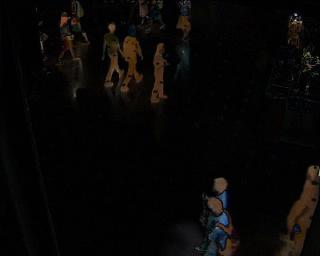}}\hfill
\subfloat{\includegraphics[width=0.16\linewidth]{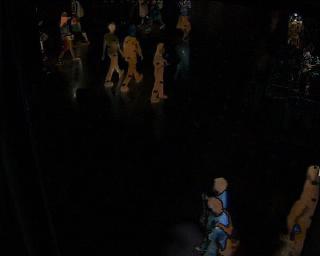}}
\\
\vspace{-0.12in}
\subfloat{\includegraphics[width=0.16\linewidth]{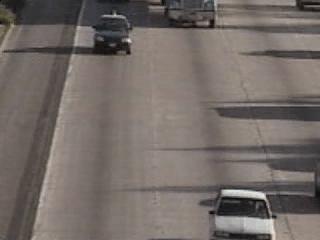}}\hfill
\subfloat{\includegraphics[width=0.16\linewidth]{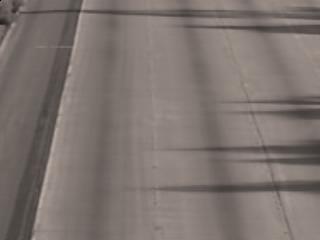}}\hfill
\subfloat{\includegraphics[width=0.16\linewidth]{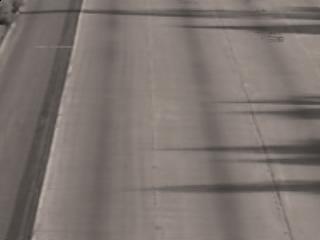}}\hfill
\subfloat{\includegraphics[width=0.16\linewidth]{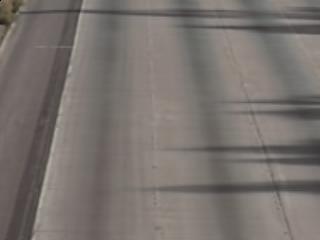}}\hfill
\subfloat{\includegraphics[width=0.16\linewidth]{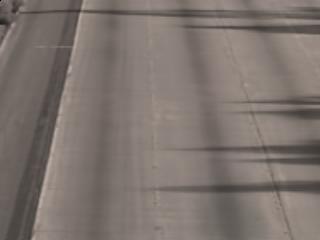}}\hfill
\subfloat{\includegraphics[width=0.16\linewidth]{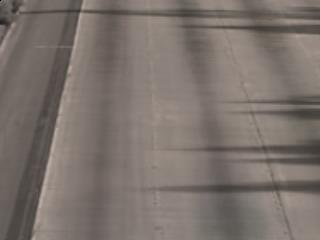}}
\\
\vspace{-0.12in}
\subfloat{\includegraphics[width=0.16\linewidth]{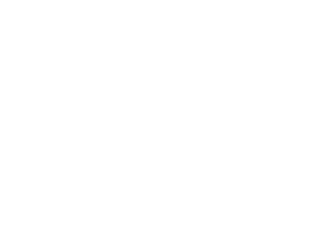}}\hfill
\subfloat{\includegraphics[width=0.16\linewidth]{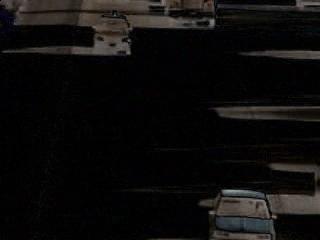}}\hfill
\subfloat{\includegraphics[width=0.16\linewidth]{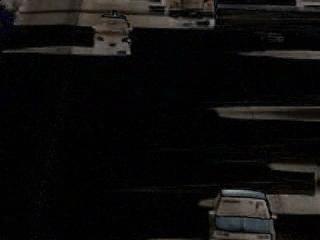}}\hfill
\subfloat{\includegraphics[width=0.16\linewidth]{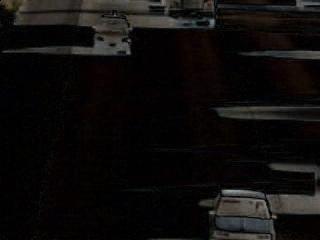}}\hfill
\subfloat{\includegraphics[width=0.16\linewidth]{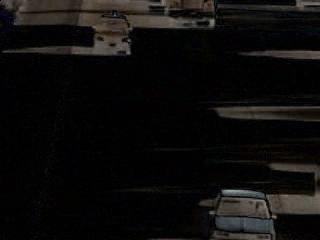}}\hfill
\subfloat{\includegraphics[width=0.16\linewidth]{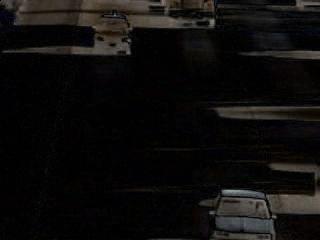}} 
\\
\vspace{-0.12in}
\subfloat{\includegraphics[width=0.16\linewidth]{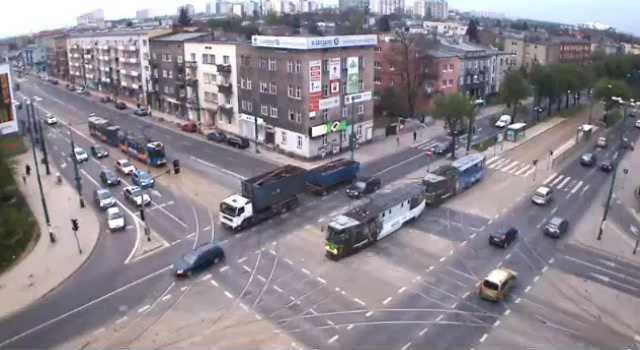}}\hfill
\subfloat{\includegraphics[width=0.16\linewidth]{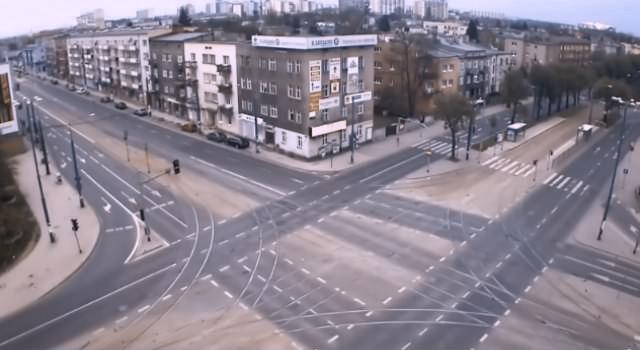}}\hfill
\subfloat{\includegraphics[width=0.16\linewidth]{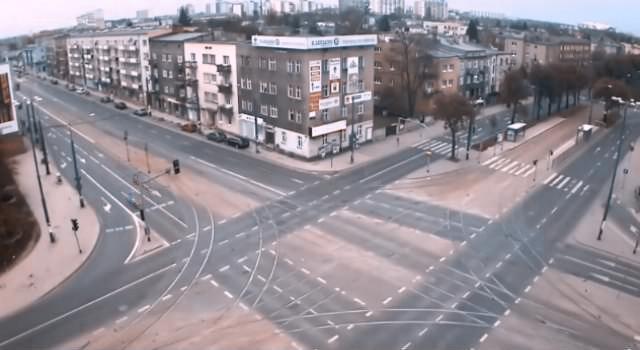}}\hfill
\subfloat{\includegraphics[width=0.16\linewidth]{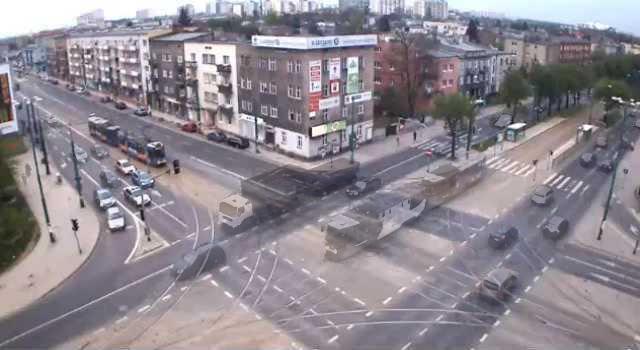}}\hfill
\subfloat{\includegraphics[width=0.16\linewidth]{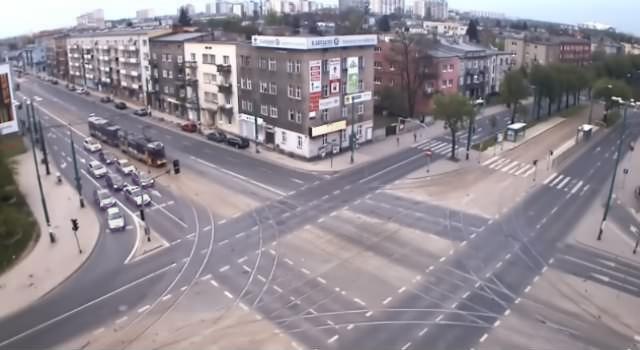}}\hfill
\subfloat{\includegraphics[width=0.16\linewidth]{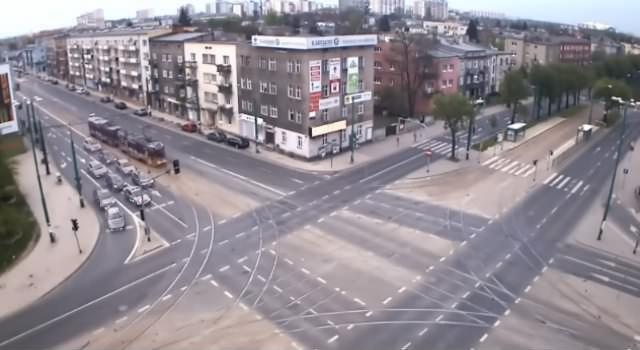}}
\vspace{-0.12in}
\\
\subfloat{\includegraphics[width=0.16\linewidth]{video2_blank.jpg}}\hfill
\subfloat{\includegraphics[width=0.16\linewidth]{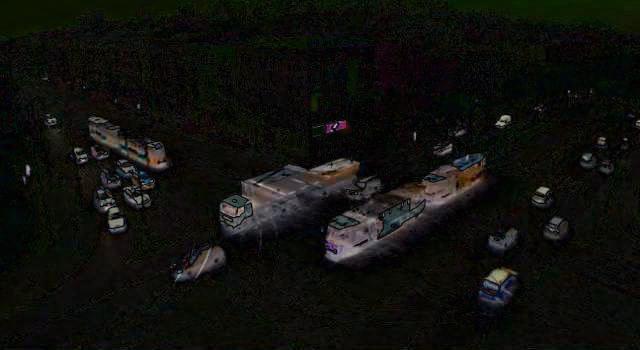}}\hfill
\subfloat{\includegraphics[width=0.16\linewidth]{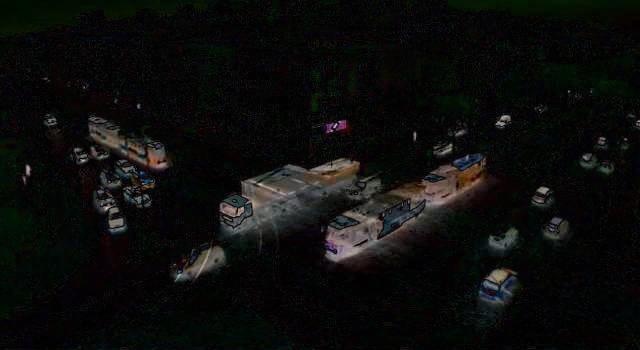}}\hfill
\subfloat{\includegraphics[width=0.16\linewidth]{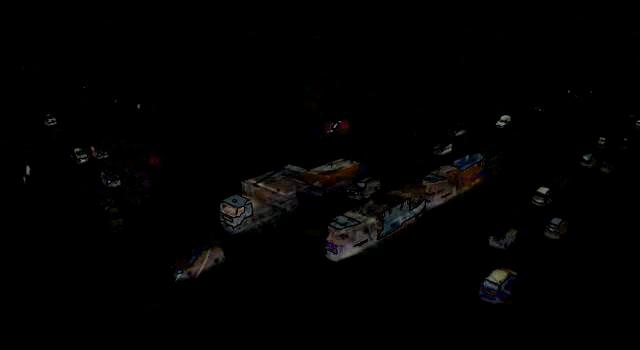}}\hfill
\subfloat{\includegraphics[width=0.16\linewidth]{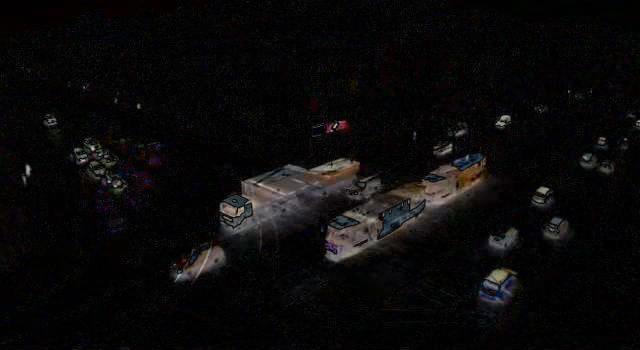}}\hfill
\subfloat{\includegraphics[width=0.16\linewidth]{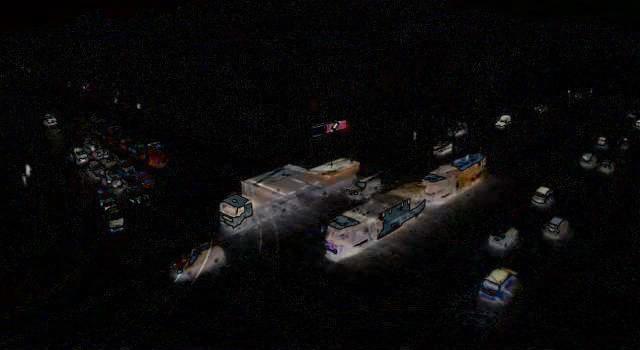}} 
\\
\vspace{-0.34in}
\subfloat{\includegraphics[width=0.16\linewidth]{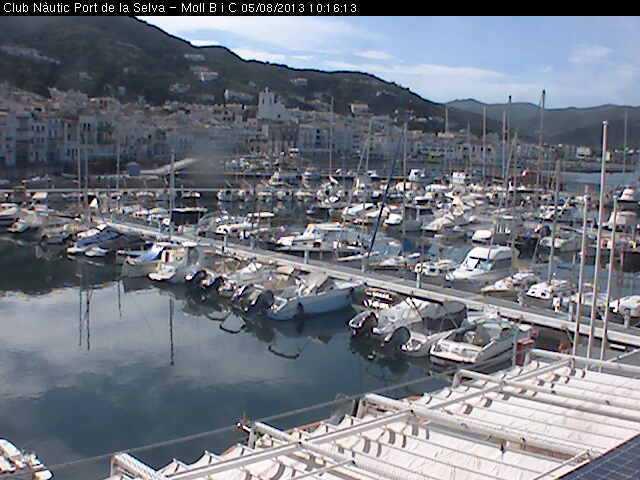}}\hfill
\subfloat{\includegraphics[width=0.16\linewidth]{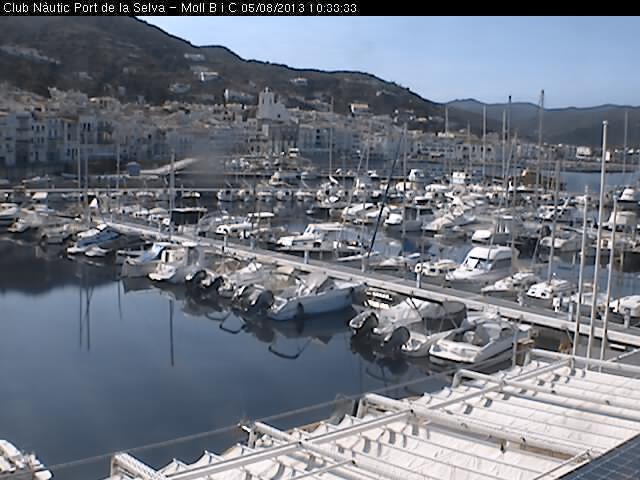}}\hfill
\subfloat{\includegraphics[width=0.16\linewidth]{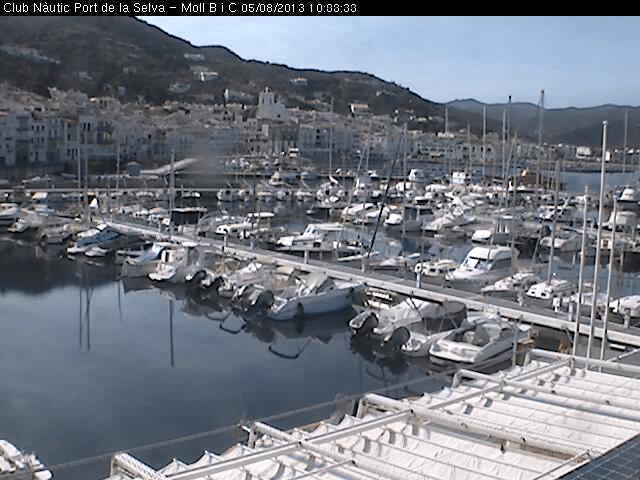}}\hfill
\subfloat{\includegraphics[width=0.16\linewidth]{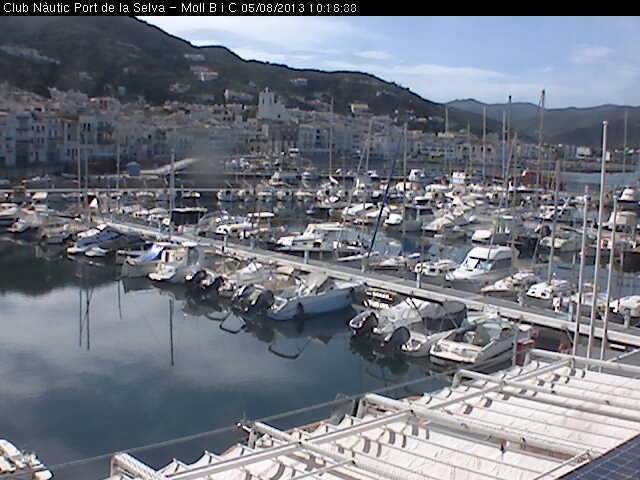}}\hfill
\subfloat{\includegraphics[width=0.16\linewidth]{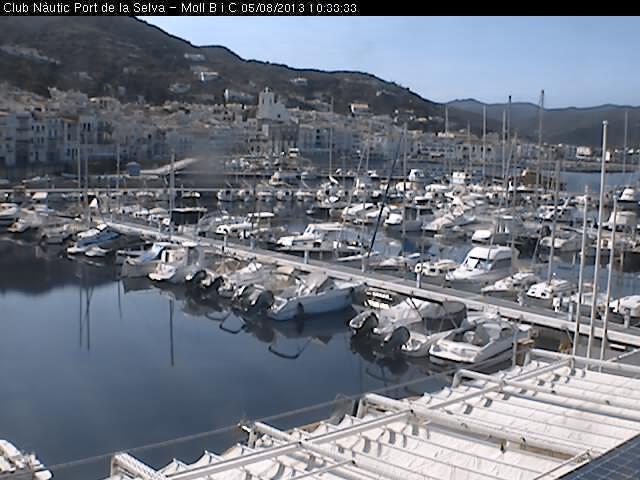}}\hfill
\subfloat{\includegraphics[width=0.16\linewidth]{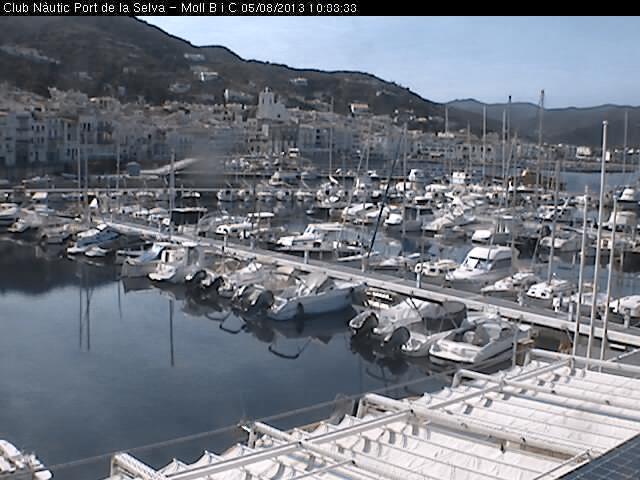}}
\vspace{-0.12in}
\\
\subfloat{\includegraphics[width=0.16\linewidth]{video2_blank.jpg}}\hfill
\subfloat{\includegraphics[width=0.16\linewidth]{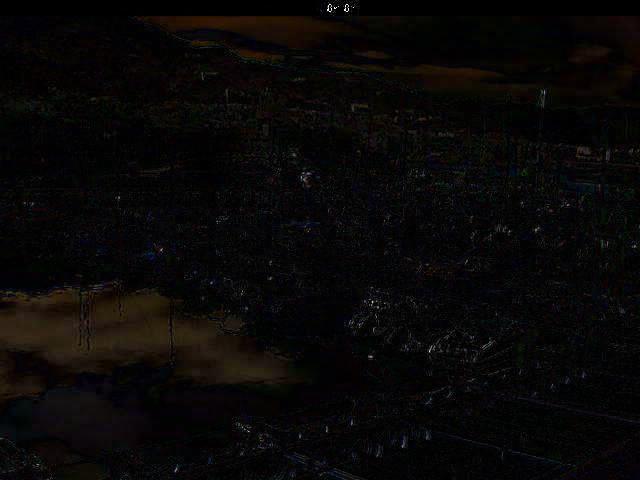}}\hfill
\subfloat{\includegraphics[width=0.16\linewidth]{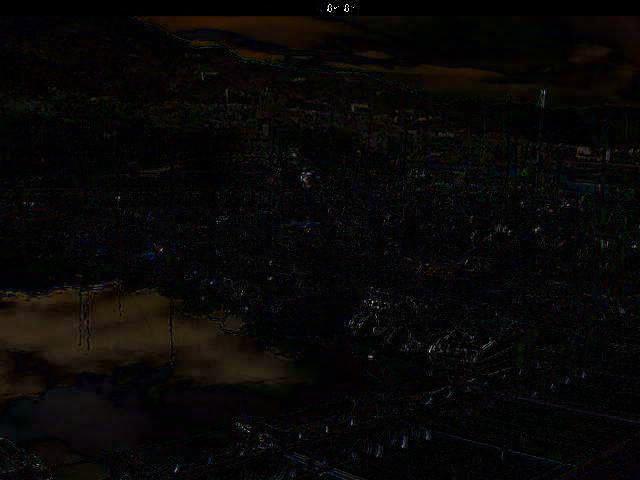}}\hfill
\subfloat{\includegraphics[width=0.16\linewidth]{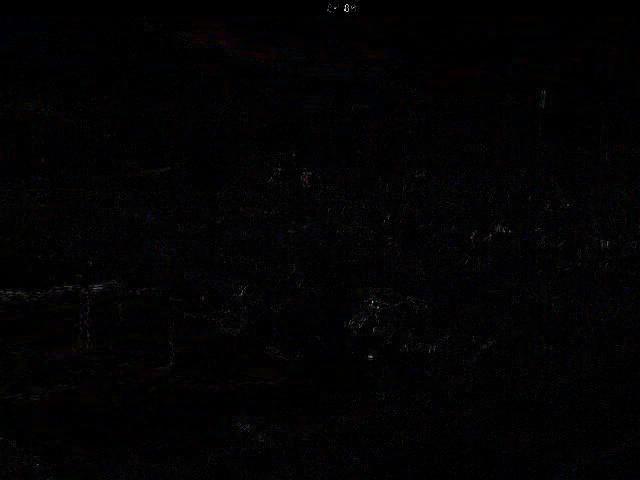}}\hfill
\subfloat{\includegraphics[width=0.16\linewidth]{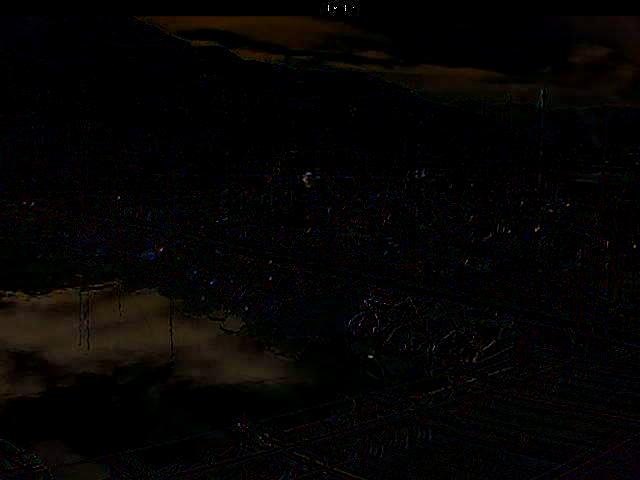}}\hfill
\subfloat{\includegraphics[width=0.16\linewidth]{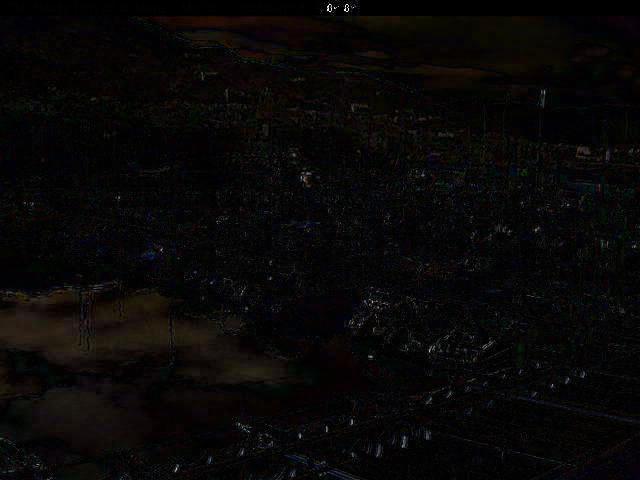}} 
\vspace{-0.12in}

\subfloat{\includegraphics[width=0.16\linewidth]{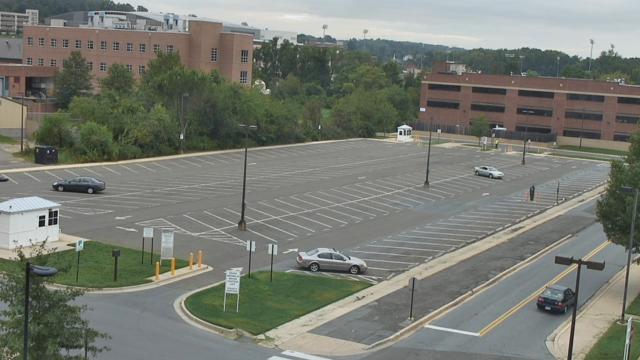}}\hfill
\subfloat{\includegraphics[width=0.16\linewidth]{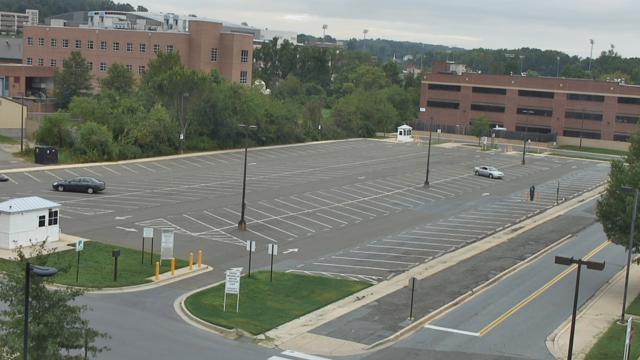}}\hfill
\subfloat{\includegraphics[width=0.16\linewidth]{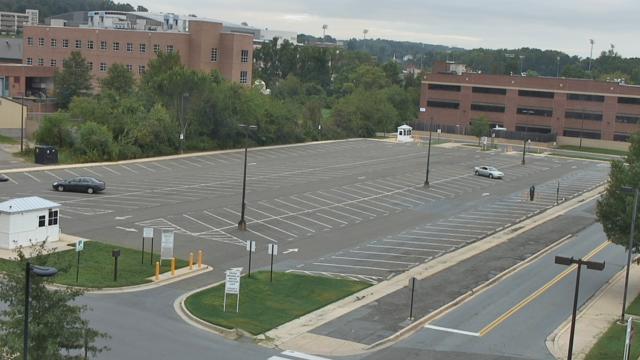}}\hfill
\subfloat{\includegraphics[width=0.16\linewidth]{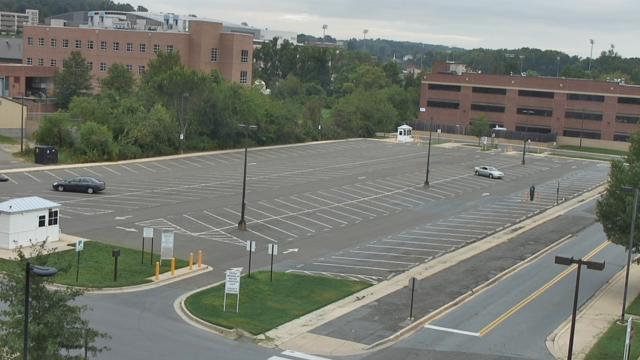}}\hfill
\subfloat{\includegraphics[width=0.16\linewidth]{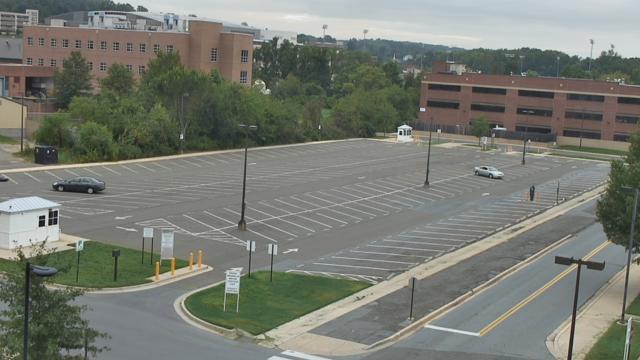}}\hfill
\subfloat{\includegraphics[width=0.16\linewidth]{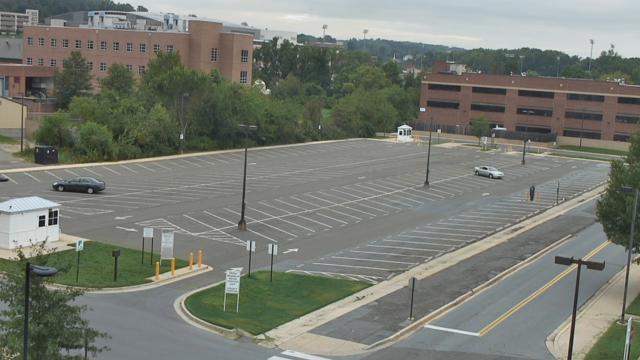}}
\vspace{-0.12in}
\\
\subfloat{\includegraphics[width=0.16\linewidth]{video2_blank.jpg}}\hfill
\subfloat{\includegraphics[width=0.16\linewidth]{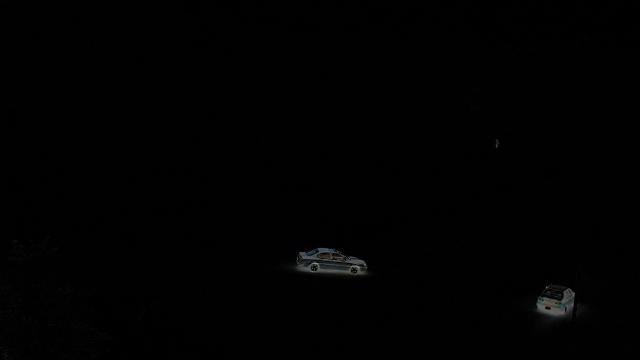}}\hfill
\subfloat{\includegraphics[width=0.16\linewidth]{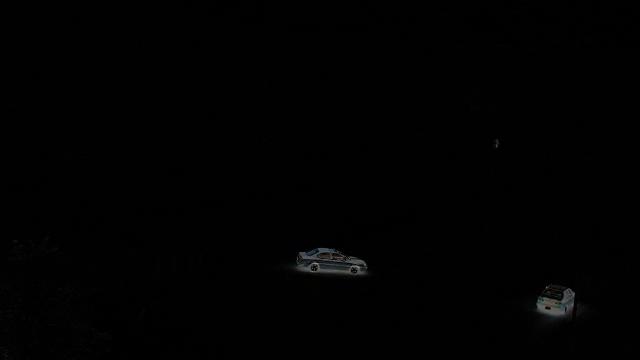}}\hfill
\subfloat{\includegraphics[width=0.16\linewidth]{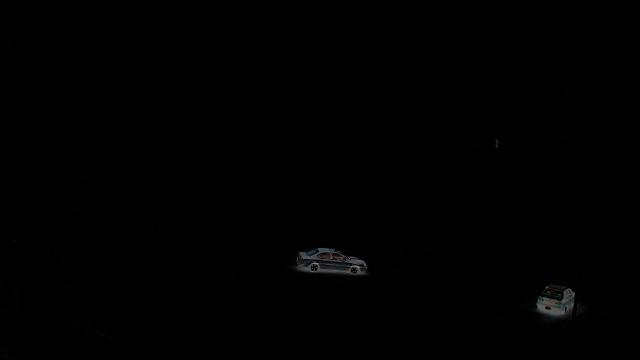}}\hfill
\subfloat{\includegraphics[width=0.16\linewidth]{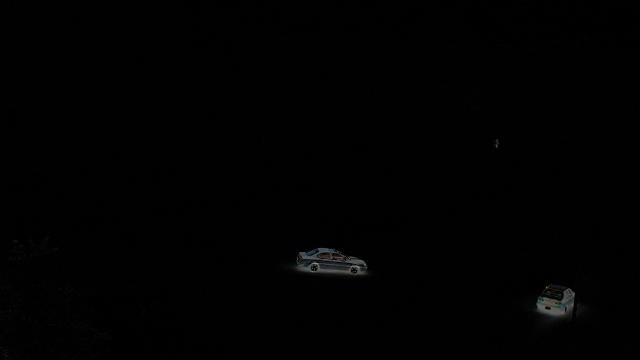}}\hfill
\subfloat{\includegraphics[width=0.16\linewidth]{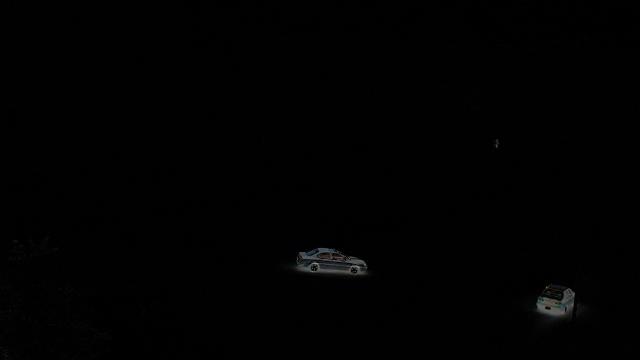}} 
\vspace{-0.25in}
\caption{Visual results for color video background subtraction. The \textbf{first two rows} are separated   backgrounds and foregrounds corresponding to a frame from \textit{Shoppingmall}, the \textbf{3rd and 4th rows}  are separated  backgrounds and foregrounds corresponding to a frame from \textit{Highway}, the \textbf{5th and 6th} are separated  backgrounds and foregrounds corresponding to a frame from \textit{Crossroad}, \textbf{7th and 8th} correspond to a frame from \textit{Port}, and the \textbf{last two rows} correspond to a frame from \textit{Parking lot}, except the first column which is the original frame. 
}\label{FIG: video}
\end{figure*}

The test results along with video size information are summarized in Table~\ref{table:video}. In addition, we provide some selected visual results in Figure~\ref{FIG: video}. All tested methods produce visually desirable backgrounds while \algf\ spends least time in the tests of all the three videos.

\section{Conclusion and Future Work}
{This paper presents a highly efficient algorithm \alg\ for large-scale TRPCA problems. \alg\ is developed by introducing a novel inexact low-multilinear-rank tensor approximation via Fiber CUR decomposition, whose structure provides a significantly reduced computational complexity. Specifically speaking, \alg\ has per iteration computational complexity as low as  $\cO(n^2dr^2\log^2(d)+n^2r^{n+1}\log^{n+1}(d))$ flops. Numerical experiments show that \alg\ is superior to other state-of-the-art tensor/matrix RPCA algorithms, on both synthetic and real-world datasets.}

There are four lines of future research work. First, it will be important to investigate a theoretical convergence guarantee of \alg. Second, extending \alg\ to the partially observed setting (i.e., when only a small random portion of the tensor is observed) is another future direction. Third, we will study the stability of \alg\ with additive dense noise since outliers often present with small dense noise in real-world applications. Last, there are exist tensor PCA studies focusing on the online tensor analysis in order to alleviate the heavy memory
cost \cite{sobral2014incremental,li2018online,bin2020gpu}, we will leave this topic to be our future work.

 \clearpage
{\small
\bibliographystyle{ieeetr}
\bibliography{reference}
}

\end{document}